\documentclass[acmlarge,authorversion]{acmart}
\AtBeginDocument{%
  \providecommand\BibTeX{{%
    \normalfont B\kern-0.5em{\scshape i\kern-0.25em b}\kern-0.8em\TeX}}}

\setcopyright{rightsretained}
\copyrightyear{2023}
\acmYear{2023}
\acmDOI{10.1145/3606705}
\acmJournal{JOCCH}

\usepackage{multirow}
\usepackage{longtable}
\usepackage{lscape}
\usepackage{afterpage}
\usepackage{subcaption}

\begin{document}

\pagestyle{fancy}
\fancyfoot{} 

\title{Optimization of Image Processing Algorithms for Character Recognition in Cultural Typewritten Documents}

\author{Mariana Dias}
\email{up201606486@up.pt}
\affiliation{%
  \institution{Faculty of Engineering of the University of Porto and INESC-TEC}
  \country{Portugal}
}
\author{Carla Teixeira Lopes}
\email{ctl@fe.up.pt}
\affiliation{%
  \institution{Faculty of Engineering of the University of Porto and INESC-TEC}
  \country{Portugal}
}


\begin{abstract}
Linked Data is used in various fields as a new way of structuring and connecting data. Cultural heritage institutions have been using linked data to improve archival descriptions and facilitate the discovery of information. Most archival records have digital representations of physical artifacts in the form of scanned images that are non-machine-readable. Optical Character Recognition (OCR) recognizes text in images and translates it into machine-encoded text. This paper evaluates the impact of image processing methods and parameter tuning in OCR applied to typewritten cultural heritage documents. The approach uses a multi-objective problem formulation to minimize Levenshtein edit distance and maximize the number of words correctly identified with a non-dominated sorting genetic algorithm (NSGA-II) to tune the methods’ parameters. Evaluation results show that parameterization by digital representation typology benefits the performance of image pre-processing algorithms in OCR. Furthermore, our findings suggest that employing image pre-processing algorithms in OCR might be more suitable for typologies where the text recognition task without pre-processing does not produce good results. In particular, Adaptive Thresholding, Bilateral Filter, and Opening are the best-performing algorithms for the theatre plays' covers, letters, and overall dataset, respectively, and should be applied before OCR to improve its performance.
\end{abstract}



\maketitle

\section{Introduction}

Cultural heritage institutions have the role of protecting, preserving, and facilitating access to cultural knowledge. In the current digital era, cultural material can be digitally preserved and remotely accessed by the public. Cultural heritage institutions have been adopting Linked Data to improve archival descriptions and information discovery~\cite{niu2016}. Linked data describes a web of data; it refers to a machine-readable structured data network connected to and from external data sets~\cite{bizer2009}. In cultural heritage, records’ metadata, artifacts, and concepts can be linked to improve data exploration~\cite{bromage2016}.

Digitalization methodologies are increasingly being applied to allow remote access to archival content. As digitalization in cultural heritage contents usually results in non-machine-readable representations, an initial conversion to a machine-readable format is needed to fully explore the potential of these representations in Linked Data implementations. This necessity happened in EPISA (Entity and Property Inference for Semantic Archives), a project that explores the use of Linked Data in the Portuguese Archives. As most of the digital representations in the Portuguese Archives are non-machine-readable, we first needed to convert them to a machine-readable format to build a tool~\cite{dias_article2022} that automatically suggests description values using the content of digital representations.

The process of recognizing handwritten or printed text in digital images and translating them into machine-readable documents is called Optical Character Recognition (OCR). The success of the OCR process depends on the quality of the scanned image, and it is common for heritage documents to suffer some degree of degradation over time. From uneven illumination to erased characters and angled digital representations, image processing methods can be applied before the text recognition phase to improve image quality and overall text extraction. Various parameters can tune image processing algorithms. Using default parameter values may yield bad performance results compared to the optimized combination of parameters. Therefore, the task of assigning values to parameters is essential.

We explore several image processing algorithms, using OpenCV, an image processing Python library with the Tesseract OCR engine. To solve the parameters’ optimization problem, we apply a non-dominated sorting genetic algorithm  (NSGA-II)  that instantiates the values of parameters of the image processing algorithms.
In the end, the algorithm generates a solution set for each pair of image processing algorithms and type of digital representation. In the evaluation stage, we compare the OCR performance using different image processing algorithms with the default and the obtained parameters.

The main contributions of this article are summarized as follows:

\begin{itemize}
  \item We analyze the effectiveness of NSGA-II for the parameterization of image processing algorithms;
  \item We compare global parameterization (i.e., considering all the digital representations regardless of their type) with a parameterization by digital representation type (e.g., letters, structured reports);
  \item We analyze the impact of image pre-processing on OCR considering global and by typology parameterization;
  \item We create and release a cultural heritage dataset~\cite{dias2022} of Portuguese typewritten digital representations from the 20th century and a dataset of its manual transcriptions~\cite{falcao2022};
  \item We disclose the source code\footnote{\url{https://github.com/feup-infolab/archmine}} of the parameterization of image processing algorithms.
\end{itemize}

The article is organized as follows. The various OCR-related tools and performance measures are explored in Section~\ref{background}, while the related research is detailed in Section~\ref{related_work}. Section~\ref{methodology} explains the proposed approach. Section~\ref{parameter_tuning} formalizes the problem of parameter tuning of image processing algorithms as an optimization problem and analyzes the effect of parameterization on the OCR performance, comparing it to default parameter values. We analyze the effect of each image processing algorithm on the OCR performance in Section~\ref{effects}, comparing it to an OCR scenario where no image pre-processing is used. The results are discussed in Section~\ref{sec:discussion}. Finally, conclusions are drawn in Section~\ref{conclusion}.

\section{Background} \label{background}

Optical Character Recognition is an active research field with constant new developments where most recent works use deep learning advances. In this section, we describe different OCR software, image processing algorithms, approaches to parameter optimization, and the evaluation of OCR systems.

\subsection{OCR Software}

There are currently four prominent OCR software applications: ABBYY FineReader, Google Cloud Vision API, Ocropy, and Tesseract. Both Tesseract and Ocropy are open-source engines, while ABBYY FineReader and Google Cloud Vision API are commercial engines.  In terms of approaches, all engines use Deep Neural Networks (DNN) to train their models, except for Ocropy which uses shallow networks. A DNN can improve accuracy by adding training data~\cite{drobac2020}, unachievable with shallow networks.

Furthermore, while Ocropy is a language-independent engine, ABBYY FineReader, Google Cloud Vision API, and Tesseract support dictionaries and language modeling. It is not easy to give a general definitive comparison of the different engines based on accuracy because the characteristics of different datasets influence the results. Engines are also affected by languages and font types. Moreover, recent versions of some engines have additional or updated training models and data that improve the performance and invalidate past research conclusions over which software is the best. In Table~\ref{tab:compare_ocr}, a comparison between the different highlighted technologies is available. One of the comparison criteria is whether the OCR software supports the Portuguese language because this work uses a dataset of Portuguese cultural heritage documents.

\begin{table*}[ht]
  \caption{Comparison of four OCR software applications.}
  \label{tab:compare_ocr}
\begin{tabular}{lllll}
\hline
\textbf{OCR Software} & \textbf{Approach} & \textbf{Platforms} & \textbf{License} & \textbf{Languages} \\ \hline
\begin{tabular}[c]{@{}l@{}}ABBYY\\ FineReader\end{tabular} & \begin{tabular}[c]{@{}l@{}}Convolutional Neural Networks\\ (CNN) and Recurrent Neural\\ Networks (RNN)~\cite{finereader2019}\end{tabular} & \begin{tabular}[c]{@{}l@{}}Mac,\\ Windows,\\ Online\end{tabular} & Proprietary & \begin{tabular}[c]{@{}l@{}}Supports Portuguese with\\ dictionary support~\cite{finereader2021}\end{tabular} \\ \hline
\begin{tabular}[c]{@{}l@{}}Google Cloud\\ Vision API\end{tabular} & \begin{tabular}[c]{@{}l@{}}CNN-based with language\\ models \cite{walker2018}\end{tabular} & Online & Proprietary & \begin{tabular}[c]{@{}l@{}}Supports Portuguese\\ through Latin scripts~\cite{walker2018}\end{tabular} \\ \hline
Ocropy & \begin{tabular}[c]{@{}l@{}}One-dimensional Long\\ Short-Term Memory (LSTM)\\ without language models~\cite{drobac2020}\end{tabular} & Linux & Apache & \begin{tabular}[c]{@{}l@{}}Supports languages that\\ use Latin scripts~\cite{ocropus}\end{tabular} \\ \hline
Tesseract & \begin{tabular}[c]{@{}l@{}}LSTM-based engine with\\ language models\end{tabular} & \begin{tabular}[c]{@{}l@{}}Mac,\\ Windows,\\ Linux\end{tabular} & Apache & \begin{tabular}[c]{@{}l@{}}Portuguese language\\ model with dictionary\\ package~\cite{weil2021}\end{tabular} \\ \hline
\end{tabular}
\end{table*}

\subsection{Image Processing}

Image processing techniques aim to visually enhance images and make them more suitable for human or machine interpretation. The strategies highlighted in this paper are image binarization and noise reduction.

Binarization consists of transforming colored pixels into black and white ones. Most recent implementations of binarization techniques are based on thresholding ~\cite{tensmeyer2020}. Thresholding is a segmentation method that sets pixels to black or white depending on whether the grey value is above or below a threshold value to evidence contrast between the background and foreground.

There are various noise reduction techniques, such as filtering and morphological transformations. Filters can be grouped into smoothing and sharpening. Morphological transformations are used to thin characters, connect broken strokes, and smooth contours~\cite{kumar2013}.

OpenCV is an open-source computer vision Python library launched in 1999 by Intel ~\cite{mohamad2015} with various image processing algorithms, specifically for binarization, image smoothing, and morphological transformations. The library offers four binarization operations: adaptive threshold, Otsu threshold, simple threshold, and triangle threshold (see Tables~\ref{tab:binarization_description} and \ref{tab:binarization_param}); four different smoothing operations: bilateral filter, Gaussian filter, homogeneous blur, and median blur (see Tables~\ref{tab:smoothing_description} and \ref{tab:smoothing_param}); seven different morphological transformation operations: black hat, closing, dilation, erosion, morphological gradient, opening, and top hat (see Tables~\ref{tab:morpho_description} and \ref{tab:morpho_param}). Tables~\ref{tab:binarization_description}, \ref{tab:smoothing_description}, and \ref{tab:morpho_description} describe the meaning of image processing algorithms from the OpenCV library. Tables~\ref{tab:binarization_param}, \ref{tab:smoothing_param}, and \ref{tab:morpho_param} describe the meaning and typology of image processing algorithms' parameters.

\begin{table}[ht]
  \caption{Description of binarization algorithms in the OpenCV library.}
  \label{tab:binarization_description}
\begin{tabular}{ll}
\hline
\textbf{Algorithm} & \textbf{Description} \\ \hline
Adaptive thresholding & Calculates the threshold value within the pixels' neighborhood area \\
Otsu thresholding & Automatically selects a threshold value based on interclass variance maximization \\
Simple thresholding & Applies the same global threshold value to every pixel \\
Triangle thresholding & \begin{tabular}[c]{@{}l@{}}Histogram-based algorithm that automatically selects a threshold value based on\\ interclass variance maximization\end{tabular} \\ \hline
\end{tabular}
\end{table}

\begin{table}[ht]
  \caption{Description of binarization algorithms' parameters in the OpenCV library.}
  \label{tab:binarization_param}
\begin{tabular}{lll}
\hline
\textbf{Parameter Name} & \textbf{Parameter Type} & \textbf{Meaning} \\ \hline
adaptiveMethod & nominal & \begin{tabular}[c]{@{}l@{}}Adaptive thresholding method. Takes the values: mean (0); \\or gaussian (1)\end{tabular} \\
blockSize & discrete & Size of a pixel’s neighborhood area \\
constant & continuous & Arbitrary constant \\
maxValue & continuous & \begin{tabular}[c]{@{}l@{}}Maximum value assigned to pixels that exceed the threshold value\end{tabular} \\
thresh & continuous & Threshold value \\
thresholdType & nominal & \begin{tabular}[c]{@{}l@{}}Thresholding type. Takes the values: threshold binary (0); \\or threshold binary inverted (1)\end{tabular} \\
type & nominal & \begin{tabular}[c]{@{}l@{}}Thresholding type. Takes the values threshold binary (0); \\threshold binary inverted (1); truncate (2); threshold to zero (3); \\or inverted (4)\end{tabular} \\ \hline
\end{tabular}%
\end{table}

\begin{table}[ht]
  \caption{Description of image smoothing algorithms in the OpenCV library.}
  \label{tab:smoothing_description}
\begin{tabular}{ll}
\hline
\textbf{Algorithm} & \textbf{Description} \\ \hline
Bilateral filter & Uses Gaussian filter, but its weight has one more component: difference in intensity \\
Gaussian blur & Blurs an image using Gaussian filter \\
Homogeneous blur & Blurs an image using a normalized box filter \\
Median blur & Blurs an image by replacing each pixel with a median of its neighborhood pixels \\ \hline
\end{tabular}
\end{table}

\begin{table}[ht]
  \caption{Description of image smoothing algorithms' parameters in the OpenCV library.}
  \label{tab:smoothing_param}
\begin{tabular}{lll}
\hline
\textbf{Parameter Name} & \textbf{Parameter Type} & \textbf{Meaning} \\ \hline
anchor & discrete & Anchor point \\
borderType & nominal & \begin{tabular}[c]{@{}l@{}}Border type. Takes the values: constant (0); replicate (1); reflect (2); \\default (3); or isolated (4)\end{tabular} \\
d & discrete & Diameter of each pixel neighborhood \\
ksize & discrete & Aperture linear/Gaussian kernel size \\
sigmaColor & continuous & Filter sigma in the colour space \\
sigmaSpace & continuous & Filter sigma in the coordinate space \\
sigmaX & discrete & Gaussian kernel standard deviation in X direction \\
sigmaY & discrete & Gaussian kernel standard deviation in Y direction \\ \hline
\end{tabular}
\end{table}

\begin{table}[ht]
  \caption{Description of morphological transformation algorithms in the OpenCV library.}
  \label{tab:morpho_description}
\begin{tabular}{ll}
\hline
\textbf{Algorithm} & \textbf{Description} \\ \hline
Black hat & Difference between input image and closing of the image \\
Closing & Dilation followed by erosion \\
Dilation & Enlarges boundaries of regions of foreground pixels \\
Erosion & Erodes boundaries of regions of foreground pixels \\
Morphological gradient & Difference between input image and opening of the image \\
Opening & Erosion followed by dilation \\
Top hat & Difference between input image and opening of the image \\ \hline
\end{tabular}
\end{table}

\begin{table}[ht]
  \caption{Description of morphological transformation algorithms' parameters in the OpenCV library.}
  \label{tab:morpho_param}
\begin{tabular}{lll}
\hline
\textbf{Parameter Name} & \textbf{Parameter Type} & \textbf{Meaning} \\ \hline
anchor & discrete & Anchor point \\
borderType & nominal & \begin{tabular}[c]{@{}l@{}}Border type. Takes the values: constant (0); replicate (1); reflect (2);\\ default (3); or isolated (4)\end{tabular} \\
iterations & discrete & Number of times technique is applied \\
kernel & continuous & \begin{tabular}[c]{@{}l@{}}Structuring element that determines the shape of a pixel\\ neighborhood\end{tabular} \\ \hline
\end{tabular}
\end{table}

\subsection{Parameter Optimization} \label{subsec:param_optimization}

To select appropriate values for the algorithms’ parameters, we can use default values or manual configurations based on literature, experience, or trial-and-error~\cite{probst2018}. OpenCV’s documentation presents some examples of implementations using recommended assigned values. For instance, in thresholding algorithms, 255 is the typical value assigned to the maxValue parameter. However, these configurations do not guarantee an optimal parameterization of the algorithms as the parameters’ values should be adjusted to the images’ state. As an alternative, parameter tuning can be done using a combinatorial search approach. According to the literature, the genetic and Hill-Climbing algorithms are the two best-tailored approaches to handle various parameters and their values~\cite{kotthoff2016}.

A genetic algorithm is a randomized search algorithm inspired by Charles Darwin’s theory of natural selection~\cite{yang2021}. It has the premise that individuals pass traits to their offspring, and the individuals with the best characteristics, tend to survive and have more offspring~\cite{yu1998}. Reproduction is the process in which the fitness function, i.e., the objective function, dictates the probability of having offspring for the next generation~\cite{roetzel2020}. In a minimization problem, lower fitness values mean a higher likelihood of offspring. Crossover is a process in which offspring are generated by combining traits from two parents, randomly selected members of the last population. Mutations are used to ensure a generation’s diversity onto the next one. The hill-climbing method uses neighbors in every iteration to evaluate a current solution candidate~\cite{pfaffe2017}. It greedily moves towards the neighbor with the highest value until there is no better neighbor. Only the genetic algorithm can manipulate nominal parameter types in a meaningful way from the two algorithms. In contrast, the hill-climbing algorithm requires a notion of neighborhood.

As some of the OpenCV image processing algorithms have nominal parameters, we use genetic algorithms. Genetic algorithms have many advantages in dealing with complex problems, and parallelism as multiple offspring that act as independent agents are ideal for parallel exploration of the search space. The main disadvantage is the difficulty in formulating the fitness function and parameters, such as the population size, crossover, and mutation rates~\cite{yang2021}. In particular, we adopt the Non-dominated Sorting Genetic Algorithm (NSGA-II)~\cite{blank2020} to solve the image processing optimization problem. NSGA-II is the standard metaheuristic for solving multi-objective optimization problems~\cite{nebro2022}. It has been applied to various search and optimization problems, such as scheduling problems~\cite{wang2020}, resource allocation problems~\cite{gao2019}, and optimal parameter selection problems~\cite{su2021}. The algorithm modifies the processes of mating and survival selection of individuals. Since not all individuals can survive, they are compared by rank with a binary tournament mating selection and selected as a solution with crowding distance that uses Manhattan Distance. Applying a ranking scheme encourages convergence and the crowing distance density estimator promotes diversity~\cite{nebro2022}.

\subsection{Evaluation of OCR systems} \label{eval}

There are two ways to evaluate OCR systems: via text or layout. A text-based evaluation is based on the comparison of the OCR text result against the ground truth. Recent research has four prominent text-based measures: Character Error Rate (CER), Word Error Rate (WER), Index/Count-based Bag of Words, and Flex Character Accuracy. In Table~\ref{tab:evaluation_ocr}, a comparison between the different measures is available. Traditionally, text-based metrics use the edit distance between the ground truth and the OCR output. That is the case for the indicators CER and WER which are also dependent on reading order. Reading order refers to a reading route for textual elements that can be sequentially ordered lists for simple layouts, such as, a book page, or more complex in the case of newspapers where there isn’t necessarily a left-to-right order~\cite{clausner2013}. Its reading dependence makes them unreliable when applied to complex page layouts since the performance is influenced by page segmentation and text detection~\cite{clausner2013}.

\begin{table*}[ht]
    \caption{Comparison of text-based OCR performance measures.}
    \label{tab:evaluation_ocr}
    \small
\begin{tabular}{llc}
\hline
\textbf{Measure} &
  \textbf{Description} &
  \textbf{Formula} \\ \hline
\begin{tabular}[c]{@{}l@{}}Character Error Rate (CER) \cite{pletschacher2015}\end{tabular} &
  \begin{tabular}[c]{@{}l@{}}Minimum number of character edition \\operations (insertions, deletions, and\\ substitutions) needed to obtain the ground\\ truth based on Levenshtein distance.\end{tabular} &
  \begin{math}\frac{C_{\text{insertions}}+C_{\text{deletions}}+C_{\text{substitutions}}}{\text{Number of characters in text}}\end{math} \\ \hline
\begin{tabular}[c]{@{}l@{}}Word Error Rate (WER) \cite{pletschacher2015}\end{tabular} &
  \begin{tabular}[c]{@{}l@{}}Minimum number of word edition \\operations (insertions, deletions, and\\ substitutions) needed to obtain the ground\\ truth based on Levenshtein distance.\end{tabular} &
  \begin{math}\frac{W_{\text{insertions}}+W_{\text{deletions}}+W_{\text{substitutions}}}{\text{Number of words in text}}\end{math} \\ \hline
\begin{tabular}[c]{@{}l@{}}Index-based Bag of Words \cite{antonacopoulos2013}\end{tabular} &
  \begin{tabular}[c]{@{}l@{}}Uses the number of words recognized at \\least once and does not account for every\\ occurrence.\end{tabular} &
  \begin{math}\frac{\text{Number of correct words}}{\text{Number of words in text}}\end{math} \\ \hline
\begin{tabular}[c]{@{}l@{}}Count-based Bag of Words \cite{antonacopoulos2013}\end{tabular} &
  \begin{tabular}[c]{@{}l@{}}Considers the correct count of the number\\ of occurrences of recognized words.\end{tabular} &
  \begin{math}\frac{\text{Number of words with correct occurrences}}{\text{Number of words in text}}\end{math} \\ \hline
\begin{tabular}[c]{@{}l@{}}Flex Character Accuracy \cite{clausner2020}\end{tabular} &
  \begin{tabular}[c]{@{}l@{}}Based on CER but adapted to not depend on\\ reading order. Edit operations are determined\\ by text alignment and line matching, where\\ non-matched lines are insertions or deletions.\end{tabular} &
  \begin{math}\frac{L_{\text{insertions}}+L_{\text{deletions}}}{\text{Number of chunks in text}}\end{math} \\ \hline
\end{tabular}
\end{table*}

In contrast, a layout-based evaluation is based on the document page’s structure. Measures analyze the document's segmentation process in different regions, classification of the areas, and reading order. The most popular ones are segmentation errors, misclassifications, and reading order errors.

There are five types of OCR error analysis from the viewpoint of misspellings: edit operations, length effects, erroneous character positions, real-word vs. non-word errors, and word boundaries~\cite{Nguyen2019}. Edit operations can be deletions, insertions, substitutions, and transpositions of characters to transform one string into another. Edit distances of 1 indicate single-error tokens, while tokens with a higher edit distance indicate multi-error tokens. There are two types of word errors: non-word and real-world errors~\cite{Tong1996}. Typically caused by the misrecognition of characters, non-word errors occur when words are converted into invalid words. In contrast, real-world errors arise when words are changed to valid words with a different meaning~\cite{Lyu2021}. Word boundary errors are caused by the wrong insertion or deletion of white spaces, which results in incorrect split errors with multiple words being merged into one or a word being split into multiple words~\cite{Nguyen2019}.

With the edit operation analysis, there are three types of OCR errors~\cite{Carrasco2014}: misspelled characters associated with the operation of substitution, spurious symbols attributed to the insertion of characters, and missing text related to the deletion of characters. An example of these errors in the context of Cultural Heritage is the transcription of a structured report from the 20th century, ``RELATÓRIO N.º 3089'' (REPORT N.º 3089). The OCR engine reads it as ``RELÁTORIO Nº 3''. The letters ``089'' are missing. Another excerpt of the transcription reads ``Enviado pela P.I.D.E.'' (Sent by P.I.D.E.), while the OCR output is ``Enviado pela P.T.D.E.''. Here, the letter ``I'' is misinterpreted as ``T''. Another excerpt reads ``permite-se aconselhar a juventude'' (allows to advise the youth), and the OCR output is ``pDermite-se aconselhar à juventude''. There is an insertion of the character ``D'' in the word ``permite-se'' and a substitution of the character ``a'' with ``à''.

\section{Related Work} \label{related_work}

Historical documents present various challenges to the OCR process. Most engines require additional training to recognize historical fonts and languages. Digital representations of historical printed records are often noisy because of the papers’ degradation due to aging, such as smears and faded text, and other printing noise like varying kerning and leading, i.e., different spaces between letters and line-break hyphenation amongst others~\cite{gupta2007}.

A popular way to improve the OCR results in historical documents is to apply an image pre-processing step to the OCR task.
In the literature, various approaches combine image processing algorithms. \citet{harraj2015} applied illumination adjustment, grayscale conversion, unsharp masking, and Otsu binarization to counteract image distortions and improve OCR accuracy. Similarly, \citet{ntogas2009} trained a model for Finnish Fraktur fonts and analyzed different combinations of pre-processing image algorithms. \citet{koistinen2017} show a process for image binarization with manual image condition classification and image processing methods. The approach has five stages: image acquisition, image preparation, filtering methods, threshold methods, and image refinement. \citet{bui2017} use a Convolution Neural Network (CNN) to automatically select image processing algorithms according to each input document image.
Most of the previously mentioned works have historical documents as their dataset. \citet{ntogas2009} used document images from the Holy Monastery of Dousiko, Meteora, Greece, while \citet{koistinen2017} used Finnish historical documents.

According to the overall results of the previously mentioned research, the use of image processing algorithms as a pre-processing step to the OCR process improves the accuracy of the engines. \citet{koistinen2017} improved word level by 27.5\% compared to FineReader 7 or 8 and 9.2\% for FineReader 11. \citet{harraj2015} registered an improvement in OCR accuracy between 2\% to 6.8\%. \citet{bui2017} succeeded in proving the performance of engines with their automatic pre-processing method selection, with Tesseract OCR having an improvement of 25\%.

As previously discussed in Section~\ref{subsec:param_optimization}, there are various approaches to the parameter tuning of image processing algorithms. \citet{lund2013} selected seven values for thresholding from the range of 0 to 255. There is an equal incremental difference between the values and an equal division of the grayscale spectrum. This experimental method reduced the error rate to 39.1\% in comparison to the original performance of the engine. To our knowledge, there is no literature on combinatorial search algorithms applied to tuning parameters of image processing algorithms in OCR.

In contrast, there are some examples of implementations of machine learning algorithms to optimize OCR accuracy. For instance, \citet{sporici2020} developed an adaptive image pre-processing method using a reinforcement learning model to minimize the edit distance between recognized text and ground truth. The technique improved the F1-score of Tesseract 4.0 from 16.3\% to 72.9\%.

Table~\ref{tab:compare_algorithms} presents the comparison of the different approaches, algorithms, and results for the discussed articles.

\begin{table*}[ht]
    \caption{Comparison of research papers related to image pre-processing in OCR or parameter tuning.}
    \label{tab:compare_algorithms}
    \resizebox{\textwidth}{!}{%
\begin{tabular}{lllll}
\hline
\textbf{Article} & \textbf{Methodology} & \textbf{\begin{tabular}[c]{@{}l@{}}Image processing\\ algorithms\end{tabular}} & \textbf{\begin{tabular}[c]{@{}l@{}}Evaluation\\ dataset\end{tabular}} & \textbf{Results} \\ \hline
\cite{ntogas2009} & \begin{tabular}[c]{@{}l@{}}Analyzed the performance of a\\ pipeline of image processing\\ algorithms according to a manual\\ image condition classification\end{tabular} & \begin{tabular}[c]{@{}l@{}}Filtering algorithms:\\ Mean, Median, Wiener,\\ Bullerworth, and Gaussian\\ Binarization algorithms:\\ Bernsen, Niblack, Otsu,\\ and Sauvola\end{tabular} & \begin{tabular}[c]{@{}l@{}}Documents from\\ the Holy Monastery\\ of Dousiko, Meteora,\\ Greece\end{tabular} & \begin{tabular}[c]{@{}l@{}}Binarization algorithms Sauvola\\ and Niblack had the best image\\ quality\end{tabular} \\ \hline
 \cite{lund2013} & \begin{tabular}[c]{@{}l@{}}Parameter tuning of thresholding\\ image processing methods by\\ selecting parameter values with\\ basis on an even distribution in\\ a range\end{tabular} & \begin{tabular}[c]{@{}l@{}}Global thresholding,\\ Otsu, and Sauvola\end{tabular} & \begin{tabular}[c]{@{}l@{}}Collection of 19th\\ Century Mormon\\ Article\end{tabular} & \begin{tabular}[c]{@{}l@{}}Reduction in error rate of 39.1\% in\\ comparison to the original\\ performance of the OCR engine\end{tabular} \\ \hline
 \cite{harraj2015} & \begin{tabular}[c]{@{}l@{}}Pipeline of image pre-processing\\ methods\end{tabular} & \begin{tabular}[c]{@{}l@{}}Illumination adjustment,\\ grayscale conversion,\\ unsharp masking, and\\ Otsu binarization\end{tabular} & Standard dataset & \begin{tabular}[c]{@{}l@{}}OCR accuracy improved between\\ 2\% to 6.8\%\end{tabular} \\ \hline
 \cite{koistinen2017} & \begin{tabular}[c]{@{}l@{}}Analyze the performance of the\\ combination of different\\ thresholding algorithms\end{tabular} & \begin{tabular}[c]{@{}l@{}}Otsu thresholding,\\ Gaussian blur, Sauvola,\\ WolfJolion, Contrast Limited\\ Adaptive Histogram Equalized\\ (CLAHE), and Linear\\ Normalization\end{tabular} & \begin{tabular}[c]{@{}l@{}}Historical Finnish\\ documents\end{tabular} & \begin{tabular}[c]{@{}l@{}}Combination of four methods\\ (Linear Normalization+WolfJolion,\\ CLAHE+WolfJolion, original image\\ and WolfJolion) had the best results\\ with an improvement of word error\\ of 1.91\% in comparison to FineReader\\ 11, and 7.21\% in comparison to\\ FineReader 7 and 8\end{tabular} \\ \hline
 \cite{bui2017} & \begin{tabular}[c]{@{}l@{}}Automatic selection of image\\ pre-processing algorithms with\\ a Convolution Neural Network\\ (CNN)\end{tabular} & \begin{tabular}[c]{@{}l@{}}Otsu binarization,\\ Su et al. binarization,\\ Gaussian filtering, non-local\\ means of noise reduction, and\\ Signh image sharpening\end{tabular} & \begin{tabular}[c]{@{}l@{}}Documents of\\ supermarket receipts\\ captured by mobile\\ devices\end{tabular} & \begin{tabular}[c]{@{}l@{}}Selection method improved\\ performance of Tesseract by 25\%.\\ The neural network had a\\ performance from 93\% to 96\%\end{tabular} \\ \hline
 \cite{sporici2020} & \begin{tabular}[c]{@{}l@{}}Reinforcement learning model\\ that minimizes edit distance for\\ an adaptive image pre-processing\\ method\end{tabular} & Adaptive thresholding & \begin{tabular}[c]{@{}l@{}}Brno Mobile OCR\\ Dataset~\cite{kiss2019}\end{tabular} & \begin{tabular}[c]{@{}l@{}}F1-score of Tesseract 4.0\\ improved from 16.3\% to 72.9\%\end{tabular} \\ \hline
\end{tabular}%
}
\end{table*}

\section{Methodology} \label{methodology}

Three research questions are associated with this experiment:

\begin{itemize}
    \item RQ1. Can applying parameter tuning to image processing algorithms with NSGA-II benefit the OCR performance of cultural heritage digital representations?
    \item RQ2. Can applying image processing algorithms to the cultural heritage digital representations before the text recognition task improve the OCR performance on the overall dataset?
    \item RQ3. Can applying image processing algorithms to the cultural heritage digital representations before the text recognition task improve the OCR performance on subsets by typology?
\end{itemize}

The methodology of this work can be divided into four main steps: define and collect a dataset of cultural heritage digital representations, explored in Section~\ref{cultural_heritage}, split the dataset into parameterization and evaluation subsets in Section~\ref{param_eval_samples}, define the image processing pipeline in Section~\ref{image_processing_pipeline}, compare the different types of parameterization and analyze the impact of image-processing in the OCR process in Section~\ref{eval_method}.

\subsection{Cultural Heritage Dataset} \label{cultural_heritage}

The training of image processing algorithms' parameters and evaluation of the OCR optimization process requires a dataset to conduct various experiments. To fit the requirements of the EPISA project, we need a Portuguese cultural heritage dataset. However, since there are currently no such available datasets, we built a dataset of digital representations of archival records. In this work, we use records from the National Archives of Torre do Tombo (ANTT), a Portuguese central archive with millions of documents that date back to the 9th century. This archive contains more than 40 million digital representations and their respective archival records~\cite{digitarq} that can be shared and accessed through an information system called Digitarq\footnote{https://digitarq.arquivos.pt/}.

To build a dataset of digital representations of archival records, we first had to identify typewritten representations in the overall set of representations. Exploring the database, we found that the document classification of the digital representations into typewritten or handwritten is usually done in the fields ``Dimension and Support'', ``Scope and Content'', and ``Note''. The description forms varied from mentions to the Portuguese word for ``typewritten'' (``datilografado''), its abbreviations (``dact.'') or alternative ways of spelling, i.e., forms of the word before and after the Portuguese Language Orthographic Agreement of 1990, ``dactilografado'', and ``datilografado''.
We searched the database for records whose descriptions in the previously mentioned fields contained different forms of the term ``typewritten''. The query results returned the name of the digital representation files and its dissemination URL that was later used to retrieve each digital representation.

The dataset~\cite{dias2022} has 27,017 one-page digital representations from 8,115 records of two fonds from the 20th century: the General Administration of National Treasury (DGFP) and the National Secretariat of Information (SNI). DGFP was a central department of the Ministry of Finance created during the First Portuguese Republic regime (1910-1926) that managed public assets, including movable and fixed assets that had been under the responsibility of the former royal family. SNI was a public agency created during the authoritarian \textit{Estado Novo} regime (1933-1974) to supervise political propaganda, press organs, popular culture, and entertainment, and promote national tourism.

\begin{figure}[ht]
    \centering
    \begin{subfigure}{.33\textwidth}
        \centering
        \includegraphics[width=\linewidth,height=5cm,keepaspectratio]{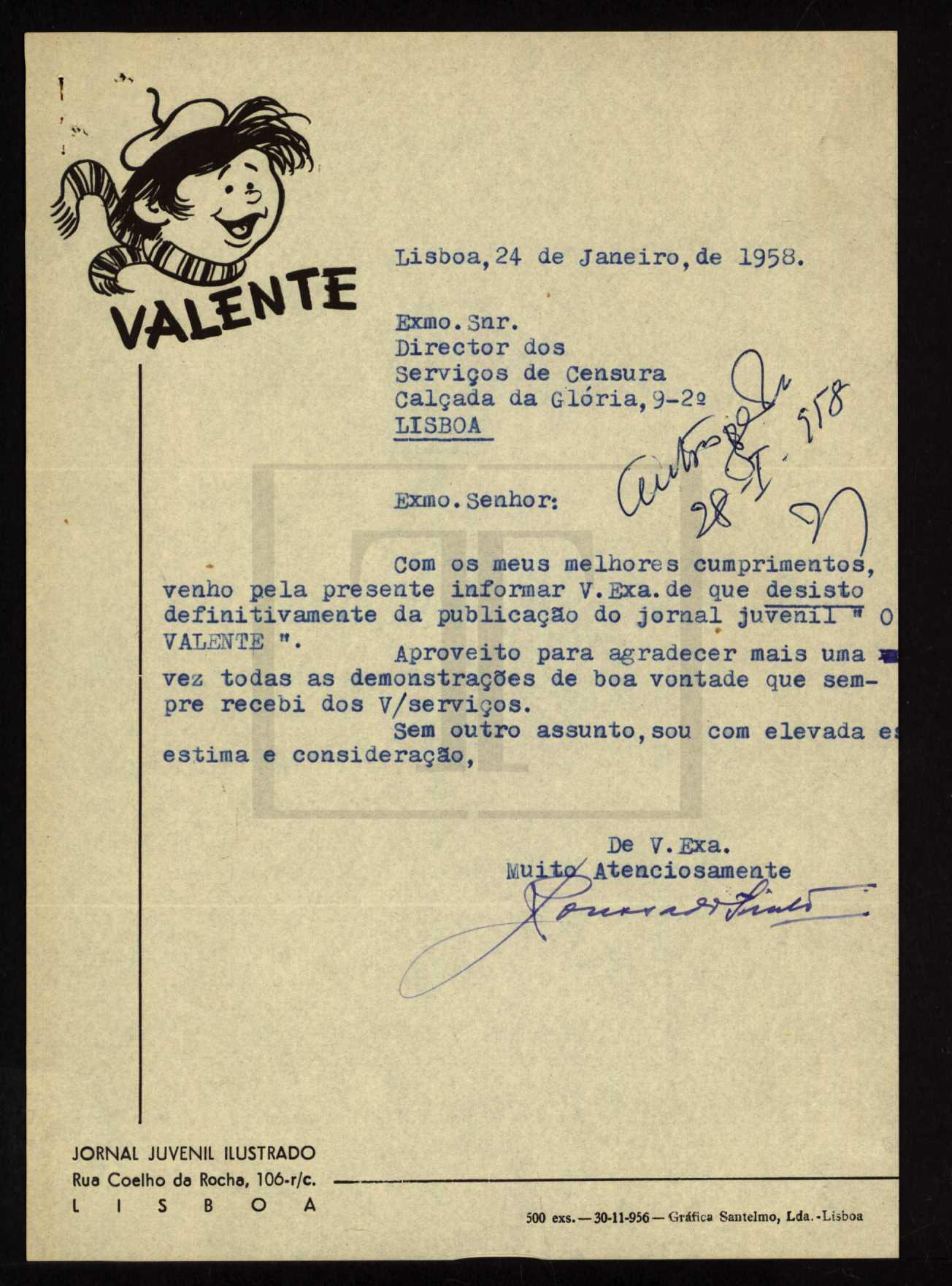}\quad
        \subcaption{Letter}
    \end{subfigure}
    \begin{subfigure}{.33\textwidth}
        \centering
        \includegraphics[width=\linewidth,height=5cm,keepaspectratio]{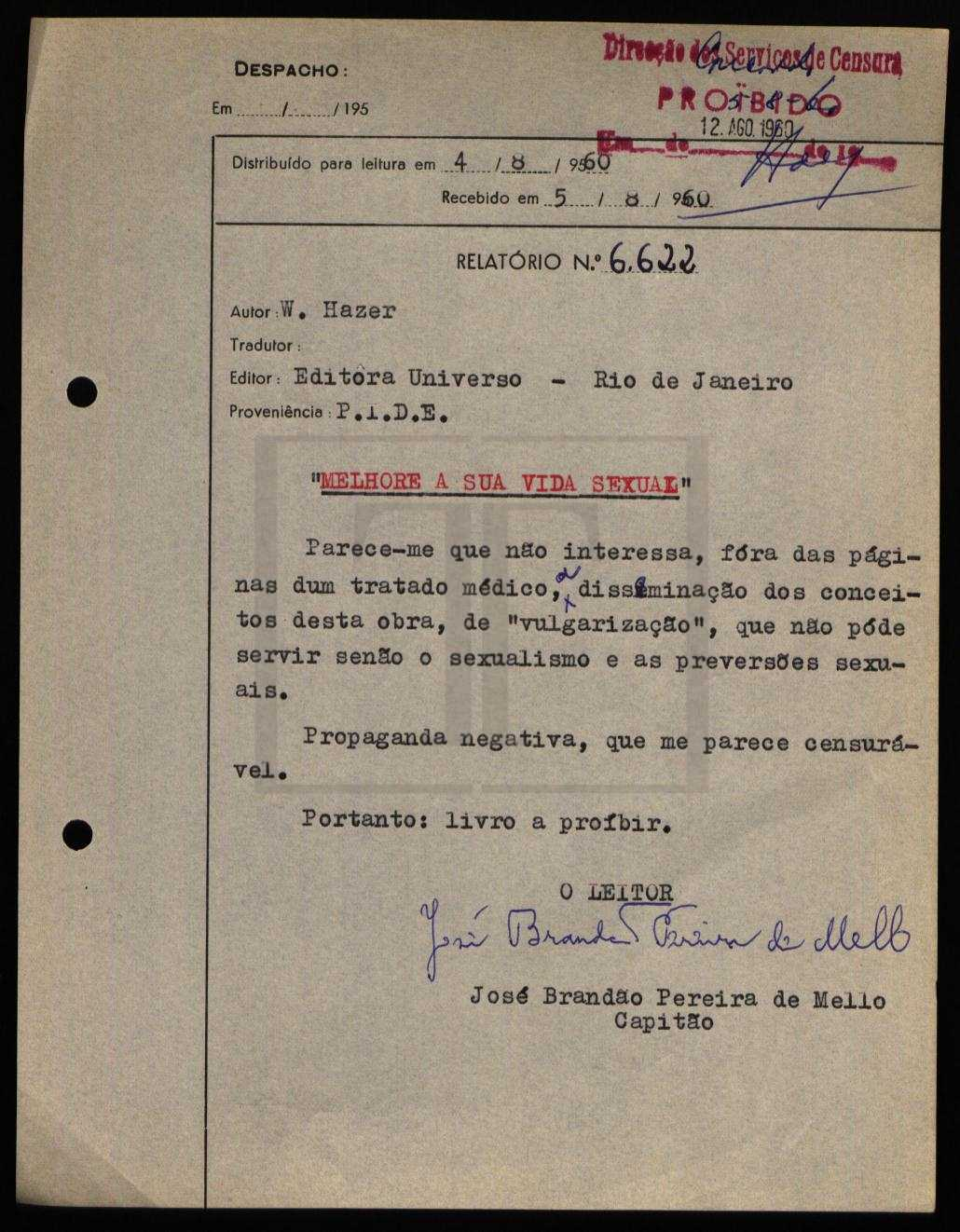}\quad
        \subcaption{Structured report}
    \end{subfigure}
    \begin{subfigure}{.33\textwidth}
        \centering
        \includegraphics[width=\linewidth,height=5cm,keepaspectratio]{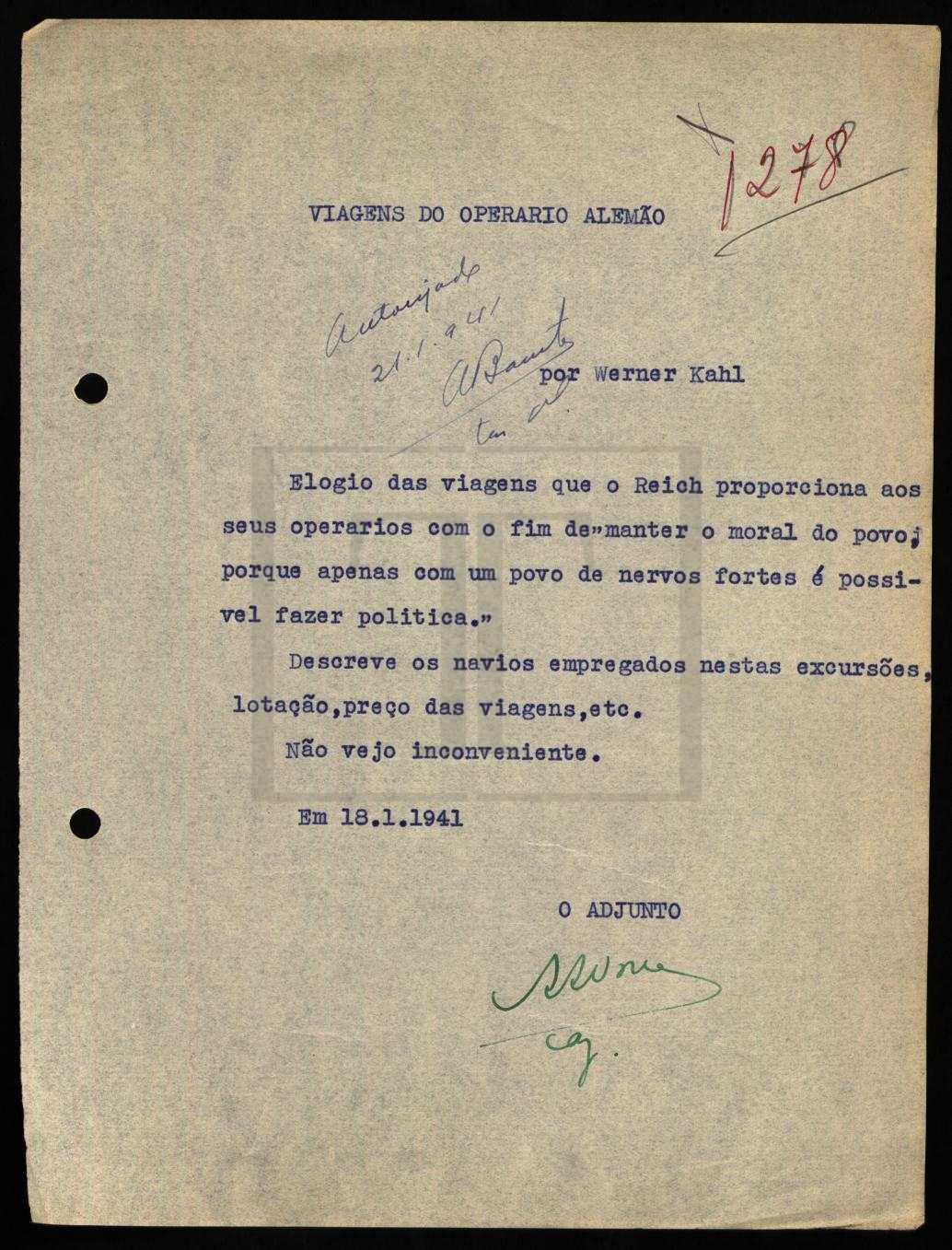}
        \subcaption{Non-structured report}
    \end{subfigure}
    \medskip
    \begin{subfigure}{.33\textwidth}
        \centering
        \includegraphics[width=\linewidth,height=5cm,keepaspectratio]{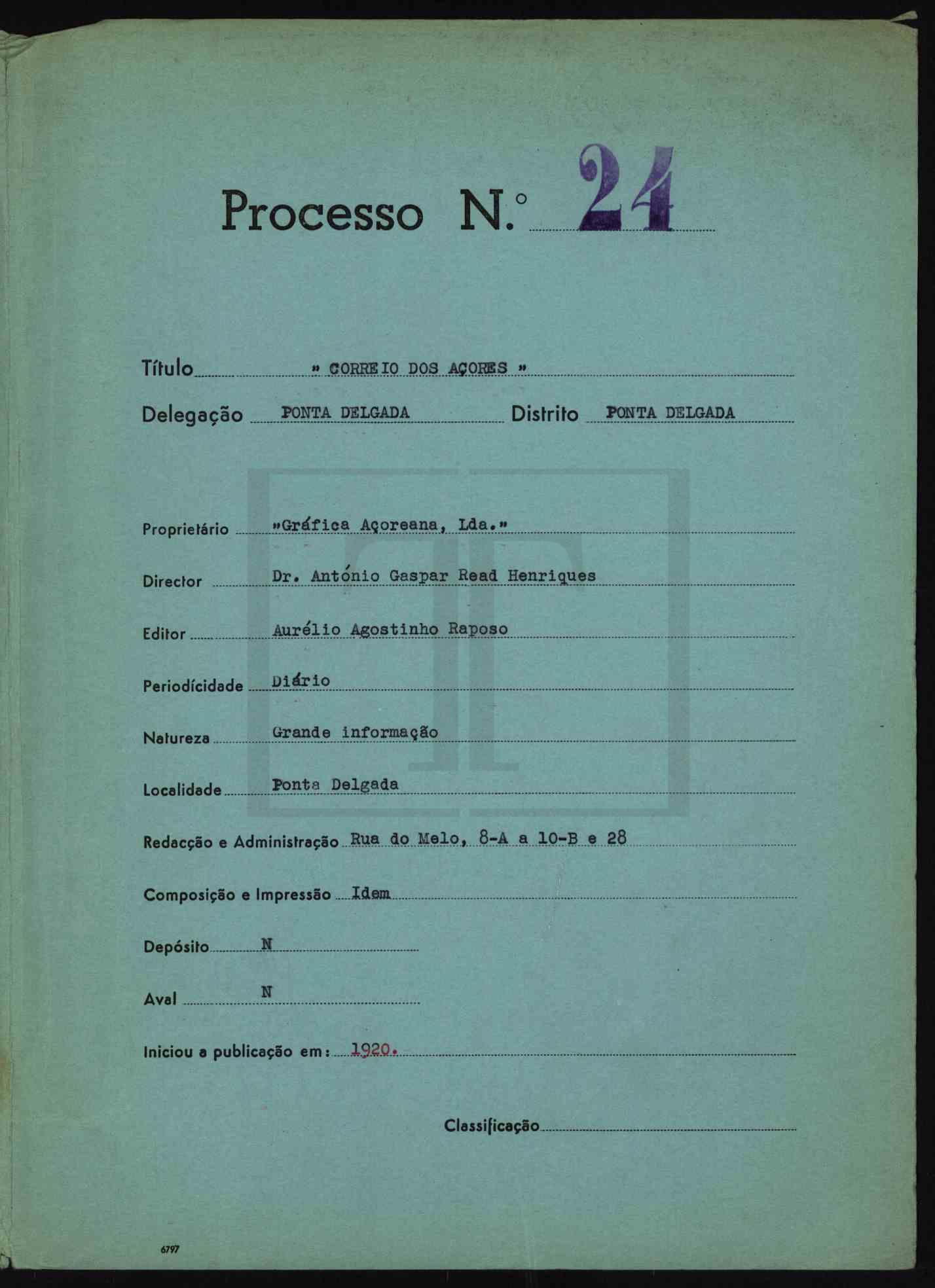}\quad
        \subcaption{Cover of a process}
    \end{subfigure}
    \begin{subfigure}{.33\textwidth}
        \centering
        \includegraphics[width=\linewidth,height=5cm,keepaspectratio]{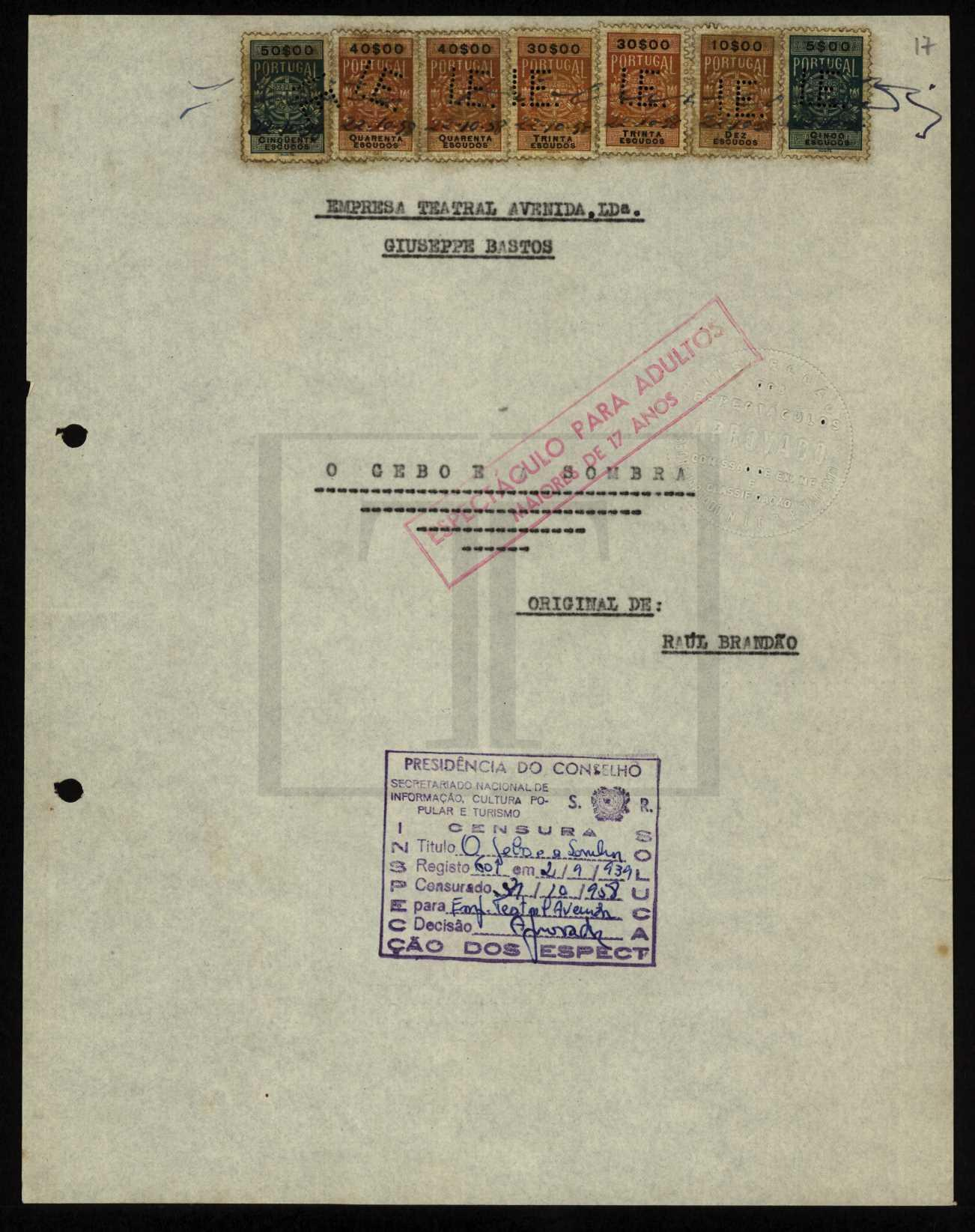}\quad
        \subcaption{Cover of a theatre play}
    \end{subfigure}
\caption{Digital representations from the typewritten dataset.}
\label{fig:dataset_example}
\end{figure}

The digital representations were classified by writing format (handwritten, typewritten, or blank), as some archival records containing handwritten or blank documents are not relevant to our work. The dataset has 23,794 typewritten digital representations, 1,681 handwritten digital representations, and 1,542 blank digital representations. Furthermore, the dataset was classified into ten types of digital representations. The typewritten dataset has 3,264 letters, 6,560 structured reports, 1,970 non-structured reports, 82 covers of processes, 6 covers of minutes, 1,473 contents of minutes, 19 books' covers, 182 books' contents, 165 theatre plays' covers, and 8,845 theatre plays' contents. Figure~\ref{fig:dataset_example} shows examples of different types of typewritten digital representations from the dataset. Identifying digital representation typologies relevant to the OCR task was carried out by observing existing textual descriptions of the records and document layouts of the digital representations. The classification of the dataset was performed manually. 

\subsection{Parameterization and Evaluation Samples} \label{param_eval_samples}

To conduct an unbiased evaluation of the OCR optimization process, we need to use different samples of the cultural heritage dataset to find the best parameter values for image processing algorithms (Parameterization Dataset) and to assess the algorithms' performance in the text recognition task (Evaluation Dataset). The best-performing image processing algorithms in the Evaluation Dataset with tuned parameters in the Parameterization Dataset will be used in the pre-processing phase of the OCR task with the Cultural Heritage Dataset to extract its textual content.

The dataset of documents we are working with refers to historical records with secondary value, i.e., evidential, and informative values, that no longer possess administrative value~\cite{dgarq2006}. Historical archives consider the documents' information as a whole, unlike current archives that detail the information contained in each document to facilitate quick access to the data. For this reason, from the ten types of digital representations of the Cultural Heritage Dataset, we decided to focus only on the five we considered most informative: letters, covers of processes, covers of theatre plays, structured and non-structured reports. These are the ones with more significant potential for describing historical records.

From the five digital representation types, we selected 708 typewritten digital representations. Per typology, we randomly selected 5\% of the digital representations of each archival series. If this methodology does not fulfill the requirement of choosing at least 60 digital representations per typology to maximize the probability of normal distributions necessary in parametric tests, we select digital representations of each archival series until the sample amounts to 60.

The digital representations of the samples were manually transcribed to provide a ground truth in the evaluation stage~\cite{falcao2022}. We split the sample into two: the Parameterization and the Evaluation Datasets. The datasets each have 81 letters, 164 structured reports, 49 non-structured reports, 30 covers of processes, and 30 theatre plays' covers. Table~\ref{tab:charact_datasets} characterizes the Parameterization and Evaluation Datasets in comparison to the Cultural Heritage Dataset regarding the number of one-page digital representations through different typologies.

\begin{table*}[ht]
    \caption{Characterization of the Parameterization, Evaluation, and Cultural Heritage Datasets' digital representations.}
    \label{tab:charact_datasets}
\resizebox{\columnwidth}{!}{%
\begin{tabular}{llccccc}
\hline
\multirow{2}{*}{\textbf{\begin{tabular}[c]{@{}l@{}}Series' Reference\\ Code\end{tabular}}} & \multirow{2}{*}{\textbf{Series Name}} & \multicolumn{5}{c}{\textbf{Parameterization / Evaluation / Cultural Heritage Datasets}} \\ \cline{3-7} 
 &  & \textbf{Letter} & \textbf{\begin{tabular}[c]{@{}c@{}}Process\\ cover\end{tabular}} & \textbf{\begin{tabular}[c]{@{}c@{}}Structured\\ report\end{tabular}} & \textbf{\begin{tabular}[c]{@{}c@{}}Theatre play\\ cover\end{tabular}} & \textbf{\begin{tabular}[c]{@{}c@{}}Non-structured\\ report\end{tabular}} \\ \hline
PT/PNA/DGFP/0001 & Royal Palace's Listings & 1/2/57 & - & - & - & - \\
PT/PNA/DGFP/0002 & \begin{tabular}[c]{@{}l@{}}List of deliveries made to\\ various institutions and\\ personalities\end{tabular} & 4/2/122 & - & - & - & - \\
PT/PNA/DGFP/0003 & \begin{tabular}[c]{@{}l@{}}Deliveries made to the Royal\\ Family and the respective\\ property appraisal processes\end{tabular} & 1/0/29 & - & - & - & - \\
PT/PNA/DGFP/0005 & \begin{tabular}[c]{@{}l@{}}Work documentation related\\ to the listings\end{tabular} & 1/0/29 & - & - & - & - \\
PT/PNA/DGFP/0006 & Correspondence & 1/5/113 & - & - & - & - \\
PT/TT/SNI-DGE/1 & \begin{tabular}[c]{@{}l@{}}Censorship cases for theatre\\ plays\end{tabular} & 7/0/138 & 18/22/54 & 0/1/20 & 30/30/165 & 0/0/1 \\
PT/TT/SNI-DGE/3 & \begin{tabular}[c]{@{}l@{}}Minutes of the Censorship\\ Commissions' Sessions\end{tabular} & - & - & - & - & - \\
PT/TT/SNI-DSC/9 & Subscription of Press Records & 16/18/699 & 8/5/17 & 0/1/6 & - & 0/2/6 \\
PT/TT/SNI-DSC/13 & \begin{tabular}[c]{@{}l@{}}Information about forbidden\\ books and period publications\end{tabular} & 41/48/1,782 & 2/1/5 & 3/0/66 & - & 0/0/2 \\
PT/TT/SNI-DSC/19 & \begin{tabular}[c]{@{}l@{}}Records on the apprehension\\ and prohibition of publications\end{tabular} & 0/2/34 & 0/1/2 & - & - & - \\
PT/TT/SNI-DSC/22 & Records on new publications & 1/0/23 & 1/0/1 & - & - & - \\
PT/TT/SNI-DSC/24 & Non-periodic publications & 0/1/22 & 1/1/3 & 0/0/1 & - & - \\
PT/TT/SNI-DSC/35 & Reports on censored books & 8/3/216 & - & 161/162/6,467 & - & 49/49/1,961 \\ \hline
Overall &  & 81/81/3,264 & 30/30/82 & 164/164/6,560 & 30/30/165 & 49/49/1,970 \\ \hline
\end{tabular}%
}
\end{table*}

\subsection{Image Processing Pipeline} \label{image_processing_pipeline}

We used the Tesseract engine for the OCR task with the Pytesseract wrapper for Python, and the available tessdata\_best~\cite{weil2021} training data. We did not train a language model, since a Portuguese language model is already available, and literature claims that the engine is prepared to handle typewritten documents.

\begin{figure}[ht]
  \centering
  \includegraphics[width=\linewidth]{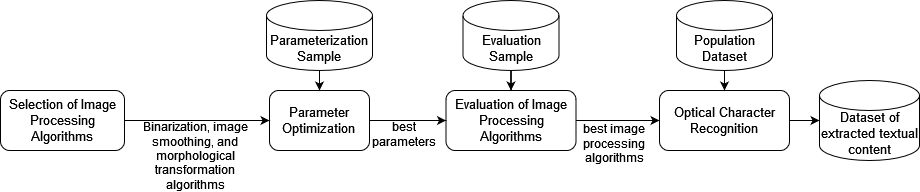}
  \caption{Overview of the different steps of the proposed methodology.}
  \label{fig:methodology}
\end{figure}

For image processing, the OpenCV library described in Table 2 was used. Three image processing techniques were selected: binarization, smoothing, and morphological. The selection of the methods was based on the state of the images in the dataset. The images’ degradation stems mainly from contrast variation, bleed-through, faded ink, smears, and thin text~\cite{sulaiman2019}, e.g., there is no angled digital representation; therefore, there is no need for a skew correction algorithm.
This work’s pipeline is shown in Figure~\ref{fig:methodology}.

\subsection{Evaluation} \label{eval_method}

Our methodology includes two evaluations that address the Research Questions defined above: the effect of parameter tuning of image processing algorithms on the OCR task (RQ1) in Section~\ref{default_effects}  and the effect of the algorithms themselves (RQ2 and RQ3) in Section~\ref{effects}.

We used performance metrics based on the CER and count-based Bag of Words presented in Table~\ref{tab:evaluation_ocr}. The CER-based metric is presented as character accuracy in Equation~\ref{eq:char_accuracy} to facilitate the comparison between the measure as success rates. The count-based Bag of Words metric is presented as an F-measure in Equation~\ref{eq:f1}, based on the notions of precision in Equation~\ref{eq:precision} and recall in Equation~\ref{eq:recall}.

\begin{equation}\label{eq:char_accuracy}
\text{Character Accuracy} = (1 - \text{CER})*100
\end{equation}

\begin{equation}\label{eq:precision}
\text{Precision}=\frac{\text{number of words with correct occurrences}}{\text{number of words in OCR output text}}
\end{equation}

\begin{equation}\label{eq:recall}
\text{Recall}=\frac{\text{number of words with correct occurrences}}{\text{number of words in ground truth text}}
\end{equation}

\begin{equation}\label{eq:f1}
\text{F1}=2*\frac{\text{precision}*\text{recall}}{\text{precision}+\text{recall}}
\end{equation}

The evaluation of the effect of image processing algorithms also includes an analysis of OCR error types. We analyzed the relation of each algorithm with OCR error types based on the OCR error analysis of edit operations described in Section~\ref{eval}.

\section{Parameter Tuning of Image Processing Algorithms} \label{parameter_tuning}

This section defines the problem of parameter tuning of image processing algorithms as an optimization problem and analyzes the effect of parameterization on the OCR performance, comparing it to default parameter values. Section~\ref{problem_formalization} details the process of formalizing and solving the problem of parameter tuning into a multi-objective optimization. We showcase the best parameter values for image processing algorithms obtained with parameterization in Section~\ref{best_parameters}, and evaluate the effect of parameterization on the OCR performance in Section~\ref{default_effects}.

\subsection{Problem Formalization and Optimization} \label{problem_formalization}

The autotuning of image processing algorithms is defined as a constrained multi-objective problem with two objective functions, \begin{math}f_{1}(x)\end{math} and \begin{math}f_{2}(x)\end{math}, formulated by Equations~\ref{eq:lev}~\cite{fontan2016} and ~\ref{eq:bow}, respectively.

\begin{equation}\label{eq:lev}
lev_{a,b}(i,j) = \left\{ \begin{array}{cl}
max(i,j) & , \ min(i,j) = 0 \\
min \left\{ \begin{array}{cl}
lev_{a,b}(i-1,j)+1 \\
lev_{a,b}(i,j-1)+1 \\
lev_{a,b}(i-1,j-1)+1
\end{array} \right. & , \ otherwise
\end{array} \right.
\end{equation}

\begin{equation}\label{eq:bow}
bow_{a,b}(i) = \left\{ \begin{array}{cl}
1 & , \ n_{a}(i) = n_{b}(i) \\
0 & , \ otherwise
\end{array} \right.
\end{equation}

The first objective function, \begin{math}f_{1}(x)\end{math}, is the Levenshtein edit distance between the ground truth and the output. The edit distance calculates the minimum number of characters’ edit operations, such as insertions, substitutions, and deletions, to transform a string \textit{a} into \textit{b} with lengths \textit{i} and \textit{j}, respectively. In contrast, the second objective function, \begin{math}f_{2}(x)\end{math}, is the number of words with the same word occurrences between the ground truth and the output, where \textit{i} is the index of matched words between strings \textit{a} and \textit{b}, and \begin{math}n_{a}(i)\end{math} is the number of occurrences of the matched word \textit{i} in string \textit{a}. The goal is to find solutions for the parameter values that satisfy each image processing algorithm’s constraints and are as good as possible for both objective functions~\cite{blank2020}.

We used the Non-dominated Sorting Genetic Algorithm (NSGA-II)~\cite{blank2020} to perform parameter tuning of image processing algorithms.
For the formalization of the constraints, \begin{math}g(x)\end{math}, the image processing algorithms’ equality constraints are transformed into inequality constraints \begin{math}g(x) \leq 0\end{math}. All the constraints needed for the selected image processing algorithms are of the type “a parameter must be odd”, which can be expressed through the equality constraint \begin{math}|\text{parameter} \% 2|=1\end{math} and transformed into the inequality \begin{math}|\text{parameter} \% 2-1| \leq 0\end{math}, where \begin{math}g(x)=|\text{parameter} \% 2-1|\end{math}. The algorithms Adaptive Thresholding, Homogeneous Blur, Median Blur, and Gaussian Blur all have a parity constraint on the parameter blockSize, in the case of the thresholding algorithm, and on the parameter ksize for the rest.

\subsection{Best Parameters Per Digital Representation Typology} \label{best_parameters}

We used the parameterization dataset to tune image processing algorithms. We considered two parameterization scenarios: global and by type of digital representation.
Global parameter tuning uses the overall parameterization sample instead of subsets of the sample by digital representation typology. The parameter values obtained using the two parametrization scenarios are presented in Table~\ref{tab:params}, along with the default values proposed by the OpenCV documentation. These values do not intend to be reference values. We decided to include them to show how parameterized values differ from the default values, how values differ between parameterization scenarios, and how values differ between types of digital representations.

\small
\setlength{\tabcolsep}{4pt}
\begin{longtable}{llccccccc}
    \caption{Parameters set for image processing algorithms with default values and values tuned by NGSA-II for global set of documents and classified types of digital representations.}
    \label{tab:params}
\\ \hline
\multirow{2}{*}{\textbf{Algorithm}} & \multirow{2}{*}{\textbf{Parameter}} &
\multirow{2}{*}{\textbf{Default}} & \multirow{2}{*}{\textbf{Global}} &
\multicolumn{5}{c}{\textbf{Type of digital representation}} \\ \cline{5-9} 
 &  &  &  &  \textbf{Letter} & \textbf{\begin{tabular}[c]{@{}c@{}}Process\\ cover\end{tabular}} & \textbf{\begin{tabular}[c]{@{}c@{}}Structured\\ report\end{tabular}} & \textbf{\begin{tabular}[c]{@{}c@{}}Theatre play \\cover\end{tabular}} & \textbf{\begin{tabular}[c]{@{}c@{}}Non-structured\\ report\end{tabular}} \\ \hline
\endfirsthead
\multicolumn{3}{c}%
{{\bfseries Table \thetable\ continued from previous page}} \\
\hline
\multirow{2}{*}{\textbf{Algorithm}} & \multirow{2}{*}{\textbf{Parameter}} &
\multirow{2}{*}{\textbf{Default}} & \multirow{2}{*}{\textbf{Global}} &
\multicolumn{5}{c}{\textbf{Type of digital representation}} \\ \cline{5-9} 
 &  &  &  &  \textbf{Letter} & \textbf{\begin{tabular}[c]{@{}c@{}}Process\\ cover\end{tabular}} & \textbf{\begin{tabular}[c]{@{}c@{}}Structured\\ report\end{tabular}} & \textbf{\begin{tabular}[c]{@{}c@{}}Theatre play \\cover\end{tabular}} & \textbf{\begin{tabular}[c]{@{}c@{}}Non-structured\\ report\end{tabular}} \\ \hline
 \endhead
\hline
\endfoot
\endlastfoot
\multirow{5}{*}{\begin{tabular}[c]{@{}l@{}}Adaptive\\ Thresholding\end{tabular}} & maxValue & 255 & 217 & 24 & 72 & 13 & 152 & 32 \\
 & adaptiveMethod & 0 or 1 & 1 & 0 & 0 & 0 & 1 & 0 \\
 & thresholdType & 0 & 0 & 0 & 0 & 0 & 1 & 0 \\
 & blockSize & 11 & 33 & 39 & 65 & 43 & 57 & 51 \\
 & c & 2 & 25 & 30 & 29 & 43 & 18 & 32 \\ \hline
\multirow{3}{*}{\begin{tabular}[c]{@{}l@{}}Bilateral\\ Filtering\end{tabular}} & d & 9 & 4 & 4 & 2 & 2 & 1 & 2 \\
 & sigmaColor & 75 & 10 & 4 & 85 & 36 & 6 & 18 \\
 & sigmaSpace & 75 & 231 & 31 & 52 & 100 & 191 & 253 \\ \hline
\multirow{3}{*}{\begin{tabular}[c]{@{}l@{}}Black\\ Hat\end{tabular}} & kernel & 5 & 10 & 6 & 229 & 139 & 30 & 38 \\
 & iterations & 1 & 7 & 10 & 7 & 9 & 2 & 1 \\
 & borderType & 3 & 4 & 4 & 2 & 3 & 1 & 1 \\ \hline
\multirow{3}{*}{Closing} & kernel & 5 & 1 & 1 & 1 & 1 & 1 & 1 \\
 & iterations & 1 & 2 & 2 & 4 & 9 & 4 & 3 \\
 & borderType & 3 & 3 & 1 & 0 & 1 & 2 & 3 \\ \hline
\multirow{3}{*}{Dilation} & kernel & 5 & 32 & - & 1 & 1 & 1 & 1 \\
 & iterations & 1 & 8 & - & 4 & 7 & 9 & 8 \\
 & borderType & 3 & 2 & - & 3 & 2 & 1 & 2 \\ \hline
\multirow{3}{*}{Erosion} & kernel & 5 & 1 & 1 & 1 & 1 & 1 & 2 \\
 & iterations & 1 & 5 & 7 & 1 & 5 & 2 & 1 \\
 & borderType & 3 & 3 & 2 & 2 & 1 & 3 & 1 \\ \hline
\pagebreak
\multirow{2}{*}{\begin{tabular}[c]{@{}l@{}}Gaussian\\ Blur\end{tabular}} & ksize & 5 & 3 & 3 & 3 & 3 & 3 & 3 \\
 & borderType & 3 & 1 & 0 & 1 & 1 & 1 & 1 \\ \hline
\multirow{2}{*}{\begin{tabular}[c]{@{}l@{}}Homogeneous\\ Blur\end{tabular}} & ksize & 5 & 1 & 1 & 3 & 1 & 3 & 3 \\
 & borderType & 3 & 1 & 2 & 3 & 1 & 0 & 3 \\ \hline
Median Blur & ksize & 5 & 3 & 3 & 3 & 5 & 3 & 3 \\ 
\multirow{3}{*}{\begin{tabular}[c]{@{}l@{}}Morphological\\ Gradient\end{tabular}} & kernel & 5 & 2 & 2 & 2 & 2 & 105 & 2 \\
 & iterations & 1 & 1 & 1 & 2 & 1 & 2 & 2 \\
 & borderType & 3 & 2 & 2 & 0 & 1 & 1 & 2 \\ \hline
\multirow{3}{*}{Opening} & kernel & 5 & 2 & 1 & 2 & 1 & 2 & 2 \\
 & iterations & 1 & 2 & 2 & 1 & 9 & 1 & 1 \\
 & borderType & 3 & 0 & 4 & 1 & 4 & 0 & 3 \\ \hline
\multirow{2}{*}{\begin{tabular}[c]{@{}l@{}}Otsu\\ Thresholding\end{tabular}} & maxValue & 255 & 199 & 24 & 251 & 92 & 34 & 14 \\
 & type & 0 & 3 & 0 & 0 & 3 & 3 & 0 \\ \hline
\multirow{3}{*}{\begin{tabular}[c]{@{}l@{}}Simple\\ Thresholding\end{tabular}} & thresh & 127 & 152 & 110 & 33 & 82 & 240 & 224 \\
 & maxValue & 255 & 45 & 4 & 10 & 221 & 3 & 42 \\
 & type & 0 & 1 & 3 & 3 & 3 & 4 & 4 \\ \hline
\multirow{3}{*}{\begin{tabular}[c]{@{}l@{}}Top\\ Hat\end{tabular}} & kernel & 5 & 90 & 51 & 217 & 233 & 139 & 204 \\
 & iterations & 1 & 5 & 7 & 5 & 2 & 5 & 2 \\
 & borderType & 3 & 2 & 0 & 0 & 2 & 2 & 4 \\ \hline
\multirow{2}{*}{\begin{tabular}[c]{@{}l@{}}Triangle\\ Thresholding\end{tabular}} & maxValue & 255 & 15 & 74 & 248 & 220 & 174 & 238 \\
 & type & 0 & 3 & 3 & 2 & 2 & 3 & 0 \\ \hline
\end{longtable}

We performed the parameterization on a PC with 3.20 GHz, AMD Ryzen 7 5800H, and 16GB RAM. Table~\ref{tab:exec_time} shows the average time per digital representation to parameterize image processing algorithms. 
The table shows that parameter tuning for each digital representation took the least time, in the range of 2 minutes, with the image processing algorithms Dilation, Erosion, and Morphological Gradient. In contrast, the parameterization for each digital representation took the longest with the algorithms Black Hat, Opening, and Otsu Thresholding, within 10 minutes.

\begin{table}[ht]
\caption{Average time to parameterize the image processing algorithms per digital representation.}
\label{tab:exec_time}
\begin{tabular}{llc}
\hline
\textbf{Technique} & \textbf{Algorithm} & \textbf{\begin{tabular}[c]{@{}c@{}}Average parameter tuning time\\ (min:s)\end{tabular}} \\ \hline
\multirow{4}{*}{Binarization} & Adaptive Thresholding & 06:46 \\
 & Otsu Thresholding & 10:52 \\
 & Simple Thresholding & 06:00 \\
 & Triangle Thresholding & 08:52 \\ \hline
\multirow{4}{*}{Image Smoothing} & Bilateral Filter & 09:11 \\
 & Gaussian Blur & 07:12 \\
 & Homogeneous Blur & 04:33 \\
 & Median Blur & 08:39 \\ \hline
\multirow{7}{*}{Morphological Transformation} & Black Hat & 10:29 \\
 & Closing & 05:31 \\
 & Dilation & 02:02 \\
 & Erosion & 02:59 \\
 & Morphological Gradient & 02:35 \\
 & Opening & 10:35 \\
 & Top Hat & 07:11 \\ \hline
\end{tabular}
\end{table}

\subsection{Effect of Parameterization on OCR Performance} \label{default_effects}

To evaluate the effect of default, global, and digital representation-type parameterization on the OCR performance, we compared the performances of image processing algorithms in OCR with default parameter values to algorithms with parameter values obtained by our two parameterization scenarios.
To compare the character accuracy and F1-score results in the three scenarios (default, global parameterization, and parameterization by typology), we use the Wilcoxon signed-rank test~\cite{noether1992} with the Bonferroni correction. We used non parametric tests because the assumptions for the t-test were not verified.
Table~\ref{tab:avg_default} presents the results of the one-sided Wilcoxon pairwise comparisons. We used this non parametric test because the assumptions for the t-test were not verified.

\begin{table}[ht]
\raggedright
    \caption{Performance comparison between image processing algorithms tuned with Wilcoxon pairwise comparisons using default, global and digital representation-type parameters.
    }
    \label{tab:avg_default}
\resizebox{\textwidth}{!}{%
\begin{tabular}{llcccccc}
\hline
\multirow{2}{*}{\textbf{Technique}} & \multirow{2}{*}{\textbf{Algorithm}} & \multicolumn{3}{c}{\textbf{Character Accuracy}} & \multicolumn{3}{c}{\textbf{F1-score}} \\ \cline{3-8} 
 &  & \textbf{\begin{tabular}[c]{@{}c@{}}Default vs\\ Global\end{tabular}} & \textbf{\begin{tabular}[c]{@{}c@{}}Default vs\\ Typology\end{tabular}} & \textbf{\begin{tabular}[c]{@{}c@{}}Global vs\\ Typology\end{tabular}} & \textbf{\begin{tabular}[c]{@{}c@{}}Default vs\\ Global\end{tabular}} & \textbf{\begin{tabular}[c]{@{}c@{}}Default vs\\ Typology\end{tabular}} & \textbf{\begin{tabular}[c]{@{}c@{}}Global vs\\ Typology\end{tabular}} \\ \hline
\multirow{5}{*}{Binarization} & \begin{tabular}[c]{@{}l@{}}Adaptive Thresholding (a)\end{tabular} & \textless{}\textless{}\textless{} & \textless{}\textless{}\textless{} & \textless{}\textless{}\textless{} & \textless{}\textless{}\textless{} & \textless{}\textless{}\textless{} & \textless{}\textless{}\textless{} \\
 & \begin{tabular}[c]{@{}l@{}}Adaptive Thresholding (b)\end{tabular} & \textless{}\textless{}\textless{} & \textless{}\textless{}\textless{} & \textless{}\textless{}\textless{} & \textless{}\textless{}\textless{} & \textless{}\textless{}\textless{} & \textless{}\textless{}\textless{} \\
 & \begin{tabular}[c]{@{}l@{}}Otsu Thresholding\end{tabular} & \textless{}\textless{}\textless{} & \textless{}\textless{}\textless{} & \textgreater{}\textgreater{}\textgreater{} & \textless{}\textless{}\textless{} & \textless{}\textless{}\textless{} & \textgreater{}\textgreater{}\textgreater{} \\
 & \begin{tabular}[c]{@{}l@{}}Simple Thresholding\end{tabular} & \textgreater{}\textgreater{}\textgreater{} & \textless{}\textless{}\textless{} & \textless{}\textless{}\textless{} & \textgreater{}\textgreater{}\textgreater{} & \textless{}\textless{}\textless{} & \textless{}\textless{}\textless{} \\
 & \begin{tabular}[c]{@{}l@{}}Triangle Thresholding\end{tabular} & 0.65 & \textless{}\textless{}\textless{} & \textless{}\textless{}\textless{} & \textless{}\textless{}\textless{} & \textless{}\textless{}\textless{} & \textless{}\textless{}\textless{} \\ \hline
\multirow{4}{*}{\begin{tabular}[c]{@{}l@{}}Image smoothing\end{tabular}} & \begin{tabular}[c]{@{}l@{}}Bilateral Filtering\end{tabular} & \textless{}\textless{}\textless{} & \textless{}\textless{}\textless{} & 0.17 & \textless{}\textless{}\textless{} & \textit{\textless{}\textless{}\textless{}} & 1 \\
 & \begin{tabular}[c]{@{}l@{}}Gaussian Blur\end{tabular} & \textless{}\textless{}\textless{} & \textless{}\textless{}\textless{} & 1 & \textless{}\textless{}\textless{} & \textless{}\textless{}\textless{} & 1 \\
 & \begin{tabular}[c]{@{}l@{}}Homogeneous Blur\end{tabular} & \textless{}\textless{}\textless{} & \textless{}\textless{}\textless{} & 0.19 & \textless{}\textless{}\textless{} & \textless{}\textless{}\textless{} & \textgreater{}\textgreater{} \\
 & \begin{tabular}[c]{@{}l@{}}Median Blur\end{tabular} & \textless{}\textless{}\textless{} & \textless{}\textless{}\textless{} & \textgreater{}\textgreater{}\textgreater{} & \textless{}\textless{}\textless{} & \textless{}\textless{}\textless{} & \textgreater{}\textgreater{}\textgreater{} \\ \hline
\multirow{7}{*}{\begin{tabular}[c]{@{}l@{}}Morphological\\ transformation\end{tabular}} & Black Hat & \textless{}\textless{}\textless{} & \textless{}\textless{}\textless{} & 0.2 & \textless{}\textless{}\textless{} & \textless{}\textless{}\textless{} & 0.24 \\
 & Closing & \textless{}\textless{}\textless{} & \textless{}\textless{}\textless{} & 0.25 & \textless{}\textless{}\textless{} & \textless{}\textless{}\textless{} & 0.26 \\
 & Dilation & \textgreater{}\textgreater{}\textgreater{} &\textless{}\textless{}\textless{} & \textless{}\textless{}\textless{} & \textgreater{}\textgreater{} & \textless{}\textless{}\textless{} & \textless{}\textless{}\textless{} \\
 & Erosion & \textless{}\textless{}\textless{} & \textless{}\textless{}\textless{} & 0.38 & \textless{}\textless{}\textless{} & \textless{}\textless{}\textless{} & 0.38 \\
 & \begin{tabular}[c]{@{}l@{}}Morphological Gradient\end{tabular} & \textless{}\textless{}\textless{} & \textless{}\textless{}\textless{} & \textgreater{} & \textless{}\textless{}\textless{} & \textless{}\textless{}\textless{} & 0.19 \\
 & Opening & \textless{}\textless{}\textless{} & \textless{}\textless{}\textless{} & \textless{}\textless{}\textless{} & \textless{}\textless{}\textless{} & \textless{}\textless{}\textless{} & \textless{}\textless{}\textless{} \\
 & Top Hat & \textless{}\textless{}\textless{} & \textless{}\textless{}\textless{} & 1 & \textless{}\textless{}\textless{} & \textless{}\textless{}\textless{} & 0.44 \\ \hline
\end{tabular}%
}
\\
\footnotesize \raggedright (a) Adaptive Thresholding using adaptiveMethod=0; \\
\footnotesize \raggedright (b) Adaptive Thresholding using adaptiveMethod=1; \\
\footnotesize \raggedright For < or >, P<0.05; for \textless{}\textless{} or \textgreater{}\textgreater{} P<0.01; for \textless{}\textless{}\textless{} or \textgreater{}\textgreater{}\textgreater{} P<0.001\\
\footnotesize \raggedright Displays the \textit{p}-values of Wilcoxon pairwise comparisons with symbol ``<'' for significantly worse performances and ``>'' for significantly better performances.\\
\end{table}

By analyzing the statistical outcomes, the OCR performance using digital representation-type parameterization on the image processing algorithms shows statistically significant improvements in both measures compared to the parameterization using default values. Similarly, the OCR performance using global-type parameterization also shows significant improvement in both measures compared to using default values, except for Simple Thresholding and Dilation where the performance is significantly worse. For most algorithms, there is either not a significant difference between global and digital representation-type parameterization or the specific parameterization performs better in both measures, except for the algorithms Otsu Thresholding, Homogeneous Blur, Median Blur, and Morphological Gradient, where it performs worse in at least one of the measures.

\section{Effect of Image Pre-Processing on OCR Performance} \label{effects}

While in the previous section we studied the effect of different types of parameterization on the OCR performance of image processing algorithms, this section analyzes the effect of each image processing algorithm on the OCR performance, comparing it to an OCR scenario where no image pre-processing is used.
Due to the significantly better results of the parameterization scenario that considered the type of digital representation reported in Section~\ref{default_effects}, we conduct the current evaluation with these parameterization values.
We use the evaluation dataset described in Section~\ref{param_eval_samples} and conduct our analysis using the overall set of documents in Section~\ref{compare_overall} and using subsets defined by the type of digital representation in Section~\ref{compare_type}.
We used the Friedman test to analyze OCR performance differences between the image processing algorithms, and the one-sided Wilcoxon signed-ranked test with the Bonferroni correction as a post-hoc test to discover which algorithms perform better in relation to each other and to not applying any.
We used non parametric tests because the assumptions for repeated measures ANOVA and t-test were not verified. We also analyze the frequency of OCR errors with different image processing algorithms on the overall evaluation dataset.

\subsection{Comparison of Image Pre-processing on the Overall Dataset} \label{compare_overall}

This section assesses the results we obtained using the complete evaluation dataset. We compare the performance of the several image pre-processing algorithms in Section~\ref{perf_analysis} and analyze the relation of each algorithm with OCR error types in Section~\ref{error_analysis}.

\subsubsection{Performance Analysis} \label{perf_analysis}

Table~\ref{tab:results_overall_dataset} presents the results of the statistical tests showing that, in the overall dataset, Bilateral Filter and Opening are the best-performing algorithms. Both algorithms significantly improve the OCR performance in terms of character accuracy, with the Opening algorithm reaching a more significant improvement. In contrast, Median Blur and Morphological Gradient are the lowest-performing algorithms in character accuracy, and Morphological Gradient is the lowest-performing algorithm in F1-score. Four algorithms worsen the character accuracy of OCR without image pre-processing: Otsu Thresholding, Triangle Thresholding, Median Blur, and Morphological Gradient. Similarly, nine algorithms worsen the OCR performance in terms of F1-score: Adaptive Thresholding, Otsu Thresholding, Simple Thresholding, Triangle Thresholding, Gaussian Blur, Homogeneous Blur, Median Blur, Morphological Gradient, and Black Hat.

\begin{table}[ht]
    \caption{Summary of Friedman test results and Wilcoxon pairwise comparisons for post-hoc analysis of OCR performance of image processing algorithms with the overall dataset. The bold values represent the image processing algorithms and average results that improved the performance of OCR without image pre-processing.}
    \label{tab:results_overall_dataset}
\resizebox{\textwidth}{!}{%
\begin{tabular}{lccllllllllllrclllllllllr}
\hline
\multirow{2}{*}{} & \multicolumn{1}{l}{\multirow{2}{*}{}} & \multicolumn{12}{c}{\textbf{Character Accuracy}} & \multicolumn{11}{c}{\textbf{F1-score}} \\ \cline{3-25} 
 & \multicolumn{1}{l}{} & \multicolumn{1}{c}{\textbf{Mean}} & \multicolumn{10}{c}{\textbf{Significantly worse results}} & \multicolumn{1}{c}{\textbf{\#}} & \multicolumn{1}{c}{\textbf{Mean}} & \multicolumn{9}{c}{\textbf{Significantly worse results}} & \multicolumn{1}{c}{\textbf{\#}} \\ \hline
\textbf{None} & N & 66.4 & OT & TT & MB & MG &  &  &  &  &  &  & 4 & 48.8 & AT & OT & ST & TT & GB & HB* & MB & MG & BH* & 9 \\ \hline
\textbf{Adaptive Thresholding} & AT & \textbf{68.9} & TT & MB & MG &  &  &  &  &  &  &  & 3 & 47.1 & TT & MB & MG &  &  &  &  &  &  & 3 \\
\textbf{Otsu Thresholding} & OT & 65.7 & TT & MB & MG &  &  &  &  &  &  &  & 3 & 45.0 & TT & MB & MG &  &  &  &  &  &  & 3 \\
\textbf{Simple Thresholding} & ST & 65.8 & TT & MB & MG &  &  &  &  &  &  &  & 3 & 47.1 & OT & TT & MB & MG &  &  &  &  &  & 4 \\
\textbf{Triangle Thresholding} & TT & 61.2 & MB & MG &  &  &  &  &  &  &  &  & 2 & 41.1 & MB & MG &  &  &  &  &  &  &  & 2 \\ \hline
\textbf{Bilateral Filter} & BF & \textbf{68.0} & \textbf{N*} & OT & ST & TT & GB & HB & MB & C* & MG &  & 9 & \textbf{49.1} & AT & OT & ST & TT & GB & HB** & MB & BH** & MG & 9 \\
\textbf{Gaussian Blur} & GB & 64.2 & TT & MB & MG &  &  &  &  &  &  &  & 3 & 45.9 & MB & MG & TT &  &  &  &  &  &  & 3 \\
\textbf{Homogeneous Blur} & HB & 64.8 & OT & TT & MB & MG &  &  &  &  &  &  & 4 & 47.0 & OT & TT & GB** & MB & MG &  &  &  &  & 5 \\
\textbf{Median Blur} & MB & 45.7 &  &  &  &  &  &  &  &  &  &  & 0 & 30.3 & MG &  &  &  &  &  &  &  &  & 1 \\ \hline
\textbf{Black Hat} & BH & \textbf{68.7} & TT & MB & MG &  &  &  &  &  &  &  & 3 & \textbf{49.0} & AT & OT & TT & MB & MG &  &  &  &  & 5 \\
\textbf{Closing} & C & \textbf{66.8} & OT & TT & GB* & MB & MG &  &  &  &  &  & 5 & 48.7 & AT & OT & ST** & TT & GB & HB* & MB & BH* & MG & 9 \\
\textbf{Erosion} & E & \textbf{67.0} & OT & TT & MB & MG &  &  &  &  &  &  & 4 & 48.7 & AT & OT & ST* & TT & GB & MB & MG &  &  & 7 \\
\textbf{Morphological Gradient} & MG & 49.3 &  &  &  &  &  &  &  &  &  &  & 0 & 30.3 &  &  &  &  &  &  &  &  &  & 0 \\
\textbf{Opening} & O & \textbf{67.8} & \textbf{N**} & OT & ST** & TT & GB** & HB & MB & C* & E* & MG & 10 & \textbf{49.2} & AT & OT & ST & TT & GB & HB & MB & BH** & MG & 9 \\
\textbf{Top Hat} & TH & \textbf{67.6} & OT* & TT & MB & MG &  &  &  &  &  &  & 4 & \textbf{48.9} & AT & OT & ST* & TT & GB** & MB & MG &  &  & 7 \\ \hline
\textbf{$\chi^2$} & \multicolumn{1}{l}{} & 905.97 & \multicolumn{1}{l}{} & \multicolumn{1}{l}{} & \multicolumn{1}{l}{} & \multicolumn{1}{l}{} & \multicolumn{1}{l}{} & \multicolumn{1}{l}{} & \multicolumn{1}{l}{} & \multicolumn{1}{l}{} & \multicolumn{1}{l}{} & \multicolumn{1}{l}{} & \multicolumn{1}{l}{} & 1170.5 & \multicolumn{1}{l}{} & \multicolumn{1}{l}{} & \multicolumn{1}{l}{} & \multicolumn{1}{l}{} & \multicolumn{1}{l}{} & \multicolumn{1}{l}{} & \multicolumn{1}{l}{} & \multicolumn{1}{l}{} & \multicolumn{1}{l}{} & \multicolumn{1}{l}{} \\
\textbf{\textit{p}-value} & \multicolumn{1}{l}{} & \textless 2.2e-16 & \multicolumn{1}{l}{} & \multicolumn{1}{l}{} & \multicolumn{1}{l}{} & \multicolumn{1}{l}{} & \multicolumn{1}{l}{} & \multicolumn{1}{l}{} & \multicolumn{1}{l}{} & \multicolumn{1}{l}{} & \multicolumn{1}{l}{} & \multicolumn{1}{l}{} & \multicolumn{1}{l}{} & \textless 2.2e-16 & \multicolumn{1}{l}{} & \multicolumn{1}{l}{} & \multicolumn{1}{l}{} & \multicolumn{1}{l}{} & \multicolumn{1}{l}{} & \multicolumn{1}{l}{} & \multicolumn{1}{l}{} & \multicolumn{1}{l}{} & \multicolumn{1}{l}{} & \multicolumn{1}{l}{} \\ \hline
\end{tabular}%
}
\\
\tiny \raggedright *P<0.05; **P<0.01
\end{table}

Looking at the results for the binarization algorithms, the Simple Thresholding algorithm is the best-performing algorithm. However, it does not improve the OCR performance.
For the smoothing algorithms' results, Bilateral Filter is the best-performing smoothing algorithm. Every smoothing algorithm except Bilateral Filter deteriorates F1-score when used to pre-process the images before OCR.
The results for the morphological transformation algorithms show that Opening is the best-performing algorithm in comparison to not applying an image pre-processing algorithm and other binarization, smoothing, and morphological transformation algorithms.

\subsubsection{Error Analysis} \label{error_analysis}

\begin{figure}[ht]
  \centering
  \includegraphics[width=\linewidth]{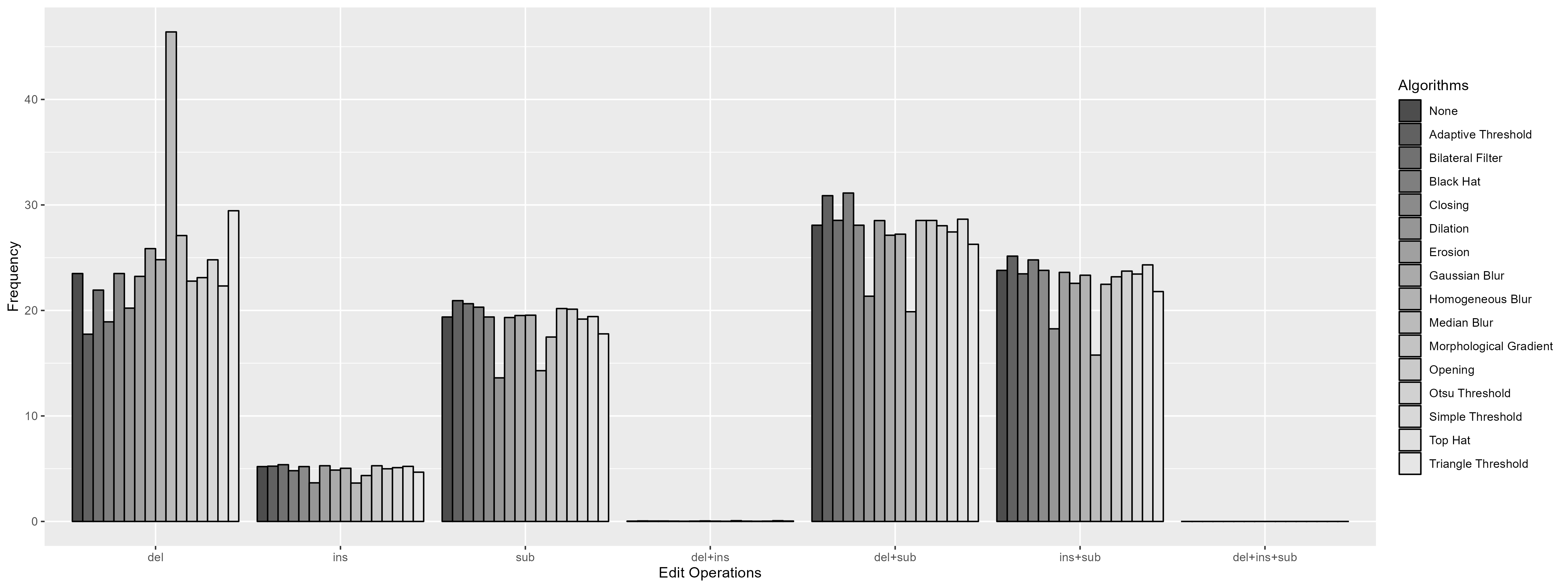}
  \caption{Frequency of edit operations for image processing algorithms in OCR. The operations of deletion, insertion and substitution are designated as del, ins and sub, and the combinations between them as del+ins, del+sub, ins+sub and del+ins+sub.}
  \label{fig:error_analysis}
\end{figure}

To better understand how image processing algorithms impact the frequency of OCR errors, we conducted an OCR error analysis of the most common errors for each image processing algorithm based on edit operation types. We present the frequency of edit operations for each image processing algorithm in Figure~\ref{fig:error_analysis}. The first graph bar in each edit operation labeled ``None'' corresponds to performing OCR without image pre-processing.

The results show the frequency of edit operations (deletion, insertion, and substitution) and combinations between them with each image processing algorithm. We conclude that among single modification error types, the operation of deletion is the most common, and insertion is the least common. Among the operation of deletion, Median Blur and Triangle Thresholding have the highest occurrences. We can also observe that the algorithms Adaptive Thresholding, Bilateral Filtering, Black Hat, Dilation, Opening, and Top Hat had a lower frequency of deletion operations in comparison to not applying any image processing algorithm to OCR. As for the combination of edit operation types where multiple errors occur, we conclude that deletion and insertion, along with deletion, insertion, and substitution have a very small occurrence with all the algorithms. The algorithms Adaptive Thresholding and Black Hat had a higher frequency of the operation of substitution combined with deletion and insertion in comparison to not using image processing algorithms.

\subsection{Comparison of Image Pre-processing on the Datasets Defined by Type of Digital Representation} \label{compare_type}

This section presents the results we obtained on algorithms comparison using subsets of the evaluation dataset defined by the type of digital representation.

\subsubsection{Letters}

This section presents the results regarding the letters' typology. Table~\ref{tab:results_letters_dataset} presents the results of the statistical tests showing that, similarly to the results for the overall dataset, Bilateral Filter is the best-performing algorithm. Triangle Thresholding and Morphological Gradient are the worst-performing algorithms. Moreover, the mean results of character accuracy and F1-score are higher than in the overall dataset.

\begin{table}[ht]
\caption{Summary of Friedman test results and Wilcoxon pairwise comparisons for post-hoc analysis of OCR performance of image processing algorithms with the letters sample. The bold values represent the image processing algorithms and average results that improved the OCR performance.}
\label{tab:results_letters_dataset}
\resizebox{\textwidth}{!}{%
\begin{tabular}{lccllllllllllrclllllllr}
\hline
\multirow{2}{*}{} & \multicolumn{1}{l}{\multirow{2}{*}{}} & \multicolumn{12}{c}{\textbf{Character Accuracy}} & \multicolumn{9}{c}{\textbf{F1-score}} \\ \cline{3-23} 
 & \multicolumn{1}{l}{} & \multicolumn{1}{c}{\textbf{Mean}} & \multicolumn{10}{c}{\textbf{Significantly worse results}} & \multicolumn{1}{c}{\textbf{\#}} & \multicolumn{1}{c}{\textbf{Mean}} & \multicolumn{7}{c}{\textbf{Significantly worse results}} & \multicolumn{1}{c}{\textbf{\#}} \\ \hline
\textbf{None} & N & 77.4 & OT & ST** & TT & MB & MG &  &  &  &  &  & 5 & 60.1 & AT & OT & ST & TT & GB** & MB & MG & 7 \\ \hline
\textbf{Adaptive Thresholding} & AT & \textbf{80.1} & TT & MB* & MG &  &  &  &  &  &  &  & 3 & 57.6 & TT & MG &  &  &  &  &  & 2 \\
\textbf{Otsu Thresholding} & OT & 75.5 & TT & MG &  &  &  &  &  &  &  &  & 2 & 52.7 & TT & MG &  &  &  &  &  & 2 \\
\textbf{Simple Thresholding} & ST & 74.4 & TT & MG &  &  &  &  &  &  &  &  & 2 & 55.3 & OT & TT & MG &  &  &  &  & 3 \\
\textbf{Triangle Thresholding} & TT & 63.3 &  &  &  &  &  &  &  &  &  &  & 0 & 37.7 &  &  &  &  &  &  &  & 0 \\ \hline
\textbf{Bilateral Filter} & BF & \textbf{78.6} & \textbf{N*} & OT & ST & TT & HB* & MB & C* & E* & MG & O* & 10 & \textbf{60.2} & AT & OT & ST & TT & GB** & MB & MG & 7 \\
\textbf{Gaussian Blur} & GB & 74.3 & TT & MB* & MG &  &  &  &  &  &  &  & 3 & 60.1 & OT & TT & MB & MG &  &  &  & 4 \\
\textbf{Homogeneous Blur} & HB & 77.4 & OT & ST** & TT & MB & MG &  &  &  &  &  & 5 & 60.1 & AT & OT & ST & TT & GB** & MB & MG & 7 \\
\textbf{Median Blur} & MB & 71.5 & MG* & TT &  &  &  &  &  &  &  &  & 2 & 54.2 & MG & TT &  &  &  &  &  & 2 \\ \hline
\textbf{Black Hat} & BH & \textbf{79.7} & TT & MB & MG &  &  &  &  &  &  &  & 3 & \textbf{60.7} & AT & OT & TT & MB** & MG &  &  & 5 \\
\textbf{Closing} & C & 77.4 & OT & ST** & TT & MB & MG &  &  &  &  &  & 5 & 60.1 & AT & OT & ST & TT & GB** & MB & MG & 7 \\
\textbf{Erosion} & E & 77.4 & OT & ST** & TT & MB & MG &  &  &  &  &  & 5 & 60.1 & AT & OT & ST & TT & GB** & MB & MG & 7 \\
\textbf{Morphological Gradient} & MG & 66.0 &  &  &  &  &  &  &  &  &  &  & 0 & 36.6 &  &  &  &  &  &  &  & 0 \\
\textbf{Opening} & O & 77.4 & OT & ST** & TT & MB & MG &  &  &  &  &  & 5 & 60.1 & AT & OT & ST & TT & GB** & MB & MG & 7 \\
\textbf{Top Hat} & TH & \textbf{77.6} & TT & MG &  &  &  &  &  &  &  &  & 2 & 59.2 & AT* & OT & TT & MB* & MG & & & 5 \\ \hline
\textbf{$\chi^2$} & \multicolumn{1}{l}{} & 223.76 & \multicolumn{1}{l}{} & \multicolumn{1}{l}{} & \multicolumn{1}{l}{} & \multicolumn{1}{l}{} & \multicolumn{1}{l}{} & \multicolumn{1}{l}{} & \multicolumn{1}{l}{} & \multicolumn{1}{l}{} & \multicolumn{1}{l}{} &  & \multicolumn{1}{l}{} & 395.56 & \multicolumn{1}{l}{} & \multicolumn{1}{l}{} & \multicolumn{1}{l}{} & \multicolumn{1}{l}{} & \multicolumn{1}{l}{} & \multicolumn{1}{l}{} & \multicolumn{1}{l}{} & \multicolumn{1}{l}{} \\
\textbf{p-value} & \multicolumn{1}{l}{} & \textless 2.2e-16 & \multicolumn{1}{l}{} & \multicolumn{1}{l}{} & \multicolumn{1}{l}{} & \multicolumn{1}{l}{} & \multicolumn{1}{l}{} & \multicolumn{1}{l}{} & \multicolumn{1}{l}{} & \multicolumn{1}{l}{} & \multicolumn{1}{l}{} &  & \multicolumn{1}{l}{} & \textless 2.2e-16 & \multicolumn{1}{l}{} & \multicolumn{1}{l}{} & \multicolumn{1}{l}{} & \multicolumn{1}{l}{} & \multicolumn{1}{l}{} & \multicolumn{1}{l}{} & \multicolumn{1}{l}{} & \multicolumn{1}{l}{} \\ \hline
\end{tabular}%
}
\\
\tiny \raggedright *P<0.05; **P<0.01
\end{table}

For the binarization methods' results, all the algorithms, except Adaptive Thresholding, decrease character accuracy. The results for F1-score are akin to the overall dataset. As for the results of the smoothing methods, Bilateral Filter is still the best-performing algorithm in character accuracy. However, the results for the morphological transformation algorithms indicate that, in comparison to the overall dataset, Opening no longer improves the OCR performance.

\subsubsection{Non-structured reports}

This section presents the results regarding the non-structured reports typology. Table~\ref{tab:results_unstructured_reports_dataset} presents the results of the statistical tests showing that the mean results of the OCR performance are slightly higher than in the overall dataset and no algorithm improves the OCR performance. Black Hat and Opening are the best-performing algorithms and Homogeneous Blur, Median Blur, and Morphological Gradient are the worst-performing algorithms.

\begin{table}[ht]
\caption{Summary of Friedman test results and Wilcoxon pairwise comparisons for post-hoc analysis of OCR performance of image processing algorithms with the non-structured reports sample. The bold values represent the average results that improved the OCR performance.}
\label{tab:results_unstructured_reports_dataset}
\resizebox{\textwidth}{!}{%
\begin{tabular}{lccllllllllrclllllllr}
\hline
 & \multicolumn{1}{l}{} & \multicolumn{10}{c}{\textbf{Character Accuracy}} & \multicolumn{9}{c}{\textbf{F1-score}} \\ \cline{3-21} 
 & \multicolumn{1}{l}{} & \multicolumn{1}{c}{\textbf{Mean}} & \multicolumn{8}{c}{\textbf{Significantly worse results}} & \multicolumn{1}{c}{\textbf{\#}} & \multicolumn{1}{c}{\textbf{Mean}} & \multicolumn{7}{c}{\textbf{Significantly worse results}} & \multicolumn{1}{c}{\textbf{\#}} \\ \hline
\textbf{None} & N & 69.2 & OT & GB & HB & MB & MG &  &  &  & 5 & 53.8 & OT & TT* & GB & HB & MB & MG &  & 6 \\ \hline
\textbf{Adaptive Thresholding} & AT & \textbf{75.8} & TT & GB* & HB* & MB** & MG &  &  &  & 5 & 53.7 & TT & MG &  &  &  &  &  & 2 \\
\textbf{Otsu Thresholding} & OT & 66.2 & GB** & HB** & MB** & MG &  &  &  &  & 4 & 44.0 & MG &  &  &  &  &  &  & 1 \\
\textbf{Simple Thresholding} & ST & 68.8 & OT & GB & HB & MB & MG &  &  &  & 5 & 53.1 & OT & TT* & GB & HB & MB & MG &  & 6 \\
\textbf{Triangle Thresholding} & TT & 67.9 & MG &  &  &  &  &  &  &  & 1 & 46.5 & MG &  &  &  &  &  &  & 1 \\ \hline
\textbf{Bilateral Filter} & BF & \textbf{69.3} & GB & HB & MB & MG &  &  &  &  & 5 & 53.8 & OT & TT* & GB & HB & MB & MG &  & 6 \\
\textbf{Gaussian Blur} & GB & 58.0 & HB* &  &  &  &  &  &  &  & 1 & 44.3 & MG &  &  &  &  &  &  & 1 \\
\textbf{Homogeneous Blur} & HB & 52.4 &  &  &  &  &  &  &  &  & 0 & 41.1 & MG &  &  &  &  &  &  & 1 \\
\textbf{Median Blur} & MB & 57.5 &  &  &  &  &  &  &  &  & 0 & 42.2 & MG &  &  &  &  &  &  & 1 \\ \hline
\textbf{Black Hat} & BH & \textbf{79.1} & OT & TT & GB & HB & MB & E* & MG & TH & 8 & \textbf{58.6} & AT** & OT & TT & GB & HB & MB & MG & 7 \\
\textbf{Closing} & C & 69.2 & OT & GB & HB & MB & MG &  &  &  & 5 & 53.8 & OT & TT* & GB & HB & MB & MG &  & 6 \\
\textbf{Dilation} & D & 69.2 & OT & GB & HB & MB & MG &  &  &  & 5 & 53.8 & OT & TT* & GB & HB & MB & MG &  & 6 \\
\textbf{Erosion} & E & \textbf{70.7} & GB** & HB** & MB** & MG &  &  &  &  & 4 & \textbf{54.1} & OT & TT** & GB* & HB** & MB & MG &  & 6 \\
\textbf{Morphological Gradient} & MG & 48.9 &  &  &  &  &  &  &  &  & 0 & 22.6 &  &  &  &  &  &  &  & 0 \\
\textbf{Opening} & O & \textbf{72.8} & OT** & TT* & BF* & GB & HB & MB & MG &  & 7 & \textbf{56.0} & OT & TT** & GB & HB & MB & MG &  & 6 \\
\textbf{Top Hat} & TH & \textbf{71.3} & GB* & HB** & MB* & MG &  &  &  &  & 4 & \textbf{55.4} & OT & TT & GB & HB & MB & MG &  & 6 \\ \hline
\textbf{$\chi^2$} & \multicolumn{1}{r}{} & 161.69 & \multicolumn{1}{l}{} & \multicolumn{1}{l}{} & \multicolumn{1}{l}{} & \multicolumn{1}{l}{} & \multicolumn{1}{l}{} & \multicolumn{1}{l}{} & \multicolumn{1}{l}{} & \multicolumn{1}{l}{} & \multicolumn{1}{l}{} & 199.79 &  &  &  &  &  &  &  &  \\
\textbf{p-value} & \multicolumn{1}{r}{} & \textless 2.2e-16 & \multicolumn{1}{l}{} & \multicolumn{1}{l}{} & \multicolumn{1}{l}{} & \multicolumn{1}{l}{} & \multicolumn{1}{l}{} & \multicolumn{1}{l}{} & \multicolumn{1}{l}{} & \multicolumn{1}{l}{} & \multicolumn{1}{l}{} & \textless 2.2e-16 & \multicolumn{1}{l}{} & \multicolumn{1}{l}{} & \multicolumn{1}{l}{} & \multicolumn{1}{l}{} & \multicolumn{1}{l}{} & \multicolumn{1}{l}{} & \multicolumn{1}{l}{} & \multicolumn{1}{l}{} \\ \hline
\end{tabular}%
}
\\
\tiny \raggedright *P<0.05; **P<0.01
\end{table}

The binarization methods Otsu and Triangle Thresholding are the worst-performing algorithms as they deteriorate the OCR performance. As for the smoothing algorithms' results, Bilateral Filter is the best-performing algorithm. All the smoothing algorithms except Bilateral Filter have significantly worse results than not using any image processing algorithm.
The results for the morphological transformation methods indicate that Black Hat and Opening are the best-performing algorithms. In sum, most morphological transformation algorithms improve the OCR performance in comparison to binarization and smoothing algorithms in the non-structured reports typology.

\subsubsection{Process covers}

This section presents the results regarding the process covers' typology. Table~\ref{tab:results_process_covers_dataset} presents the results of the statistical tests showing that, in the process covers sample, the mean OCR performance values are lower than in the overall dataset, and no algorithm improved the OCR performance.

\begin{table}[ht]
\caption{Summary of Friedman test results and Wilcoxon pairwise comparisons for post-hoc analysis of OCR performance of image processing algorithms with the process covers sample. The bold values represent the average results that improved the OCR performance.}
\label{tab:results_process_covers_dataset}
\resizebox{\textwidth}{!}{%
\begin{tabular}{lccllllrcllllr}
\hline
\multirow{2}{*}{} & \multicolumn{1}{l}{\multirow{2}{*}{}} & \multicolumn{6}{c}{\textbf{Character Accuracy}} & \multicolumn{6}{c}{\textbf{F1-score}} \\ \cline{3-14} 
 & \multicolumn{1}{l}{} & \multicolumn{1}{c}{\textbf{Mean}} & \multicolumn{4}{c}{\textbf{Significantly worse results}} & \multicolumn{1}{c}{\textbf{\#}} & \multicolumn{1}{c}{\textbf{Mean}} & \multicolumn{4}{c}{\textbf{Significantly worse results}} & \multicolumn{1}{c}{\textbf{\#}} \\ \hline
\textbf{None} & N & 62.4 & MG &  &  &  & 1 & 35.3 & BH** & MG &  &  & 2 \\ \hline
\textbf{Adaptive Thresholding} & AT & \textbf{63.1} & MG &  &  &  & 1 & 33.0 & MG &  &  &  & 1 \\
\textbf{Otsu Thresholding} & OT & 61.6 & MG &  &  &  & 1 & 31.5 & MG &  &  &  & 1 \\
\textbf{Simple Thresholding} & ST & \textbf{64.5} & BH* & MG &  &  & 2 & \textbf{36.5} & BH** & MG &  &  & 2 \\
\textbf{Triangle Thresholding} & TT & 56.3 & MG* &  &  &  & 1 & 32.8 & MG &  &  &  & 1 \\ \hline
\textbf{Bilateral Filter} & BF & \textbf{67.1} & BH** & MG &  &  & 2 & \textbf{38.9} & OT** & BH & MG &  & 3 \\
\textbf{Gaussian Blur} & GB & \textbf{67.1} & OT** & BH & MG & TH* & 4 & \textbf{39.2} & OT** & TT* & BH** & MG & 4 \\
\textbf{Homogeneous Blur} & HB & \textbf{67.4} & OT* & TT* & BH & MG & 4 & \textbf{38.8} & OT** & BH** & MG & & 3 \\
\textbf{Median Blur} & MB & \textbf{65.8} & BH* & MG &  &  & 2 & \textbf{39.3} & OT & BH** & MG &  & 3 \\ \hline
\textbf{Black Hat} & BH & 54.6 & MG &  &  &  & 1 & 28.0 & MG &  &  &  & 1 \\
\textbf{Closing} & C & 62.4 & MG &  &  &  & 1 & 35.3 & BH** & MG &  &  & 2 \\
\textbf{Dilation} & D & 62.4 & MG &  &  &  & 1 & 35.3 &  &  &  &  & 0 \\
\textbf{Erosion} & E & 62.4 & MG &  &  &  & 1 & 35.3 & BH** & MG &  &  & 2 \\
\textbf{Morphological Gradient} & MG & 34.9 &  &  &  &  & 0 & 11.4 & & & &  & 0 \\
\textbf{Opening} & O & \textbf{65.1} & BH** & MG &  &  & 2 & \textbf{36.1} & OT* & BH** & MG &  & 3 \\
\textbf{Top Hat} & TH & \textbf{62.7} & MG &  &  &  & 1 & \textbf{35.6} & MG &  &  &  & 1 \\ \hline
\textbf{$\chi^2$} & \multicolumn{1}{l}{} & 104.27 & \multicolumn{1}{l}{} & \multicolumn{1}{l}{} & \multicolumn{1}{l}{} & \multicolumn{1}{l}{} & \multicolumn{1}{l}{} & 126.29 &  &  &  &  &  \\
\textbf{p-value} & \multicolumn{1}{l}{} & 7.183e-16 & \multicolumn{1}{l}{} & \multicolumn{1}{l}{} & \multicolumn{1}{l}{} & \multicolumn{1}{l}{} & \multicolumn{1}{l}{} & \textless 2.2e-16 &  &  &  &  &  \\ \hline
\end{tabular}%
}
\\
\tiny \raggedright *P<0.05; **P<0.01
\end{table}

There are no significant differences between the binarization algorithms. Moreover, unlike in the previous sections, binarization algorithms do not impair the OCR performance.
As for the smoothing methods, we can observe that most algorithms have a better performance in character accuracy and F1-score than Black Hat, Morphological Gradient, and Otsu Thresholding.
Finally, the results with morphological transformation algorithms indicate that Black Hat and Morphological Gradient are the lowest-performing algorithms and deteriorate the OCR performance.

\subsubsection{Structured reports}

This section presents the results regarding the structured reports typology. Table~\ref{tab:results_structured_reports_dataset} presents the results of the statistical tests showing that no algorithm improved the OCR performance in comparison to not using any image processing algorithm.

\begin{table}[ht]
\caption{Summary of Friedman test results and Wilcoxon pairwise comparisons for post-hoc analysis of OCR performance of image processing algorithms with the structured reports sample. The bold values represent the average results that improved the OCR performance.}
\label{tab:results_structured_reports_dataset}
\resizebox{\textwidth}{!}{%
\begin{tabular}{lccllllllrclllllr}
\hline
\multirow{2}{*}{} & \multicolumn{1}{l}{\multirow{2}{*}{}} & \multicolumn{8}{c}{\textbf{Character Accuracy}} & \multicolumn{7}{c}{\textbf{F1-score}} \\ \cline{3-17} 
 & \multicolumn{1}{l}{} & \multicolumn{1}{c}{\textbf{Mean}} & \multicolumn{6}{c}{\textbf{Significantly worse results}} & \multicolumn{1}{c}{\textbf{\#}} & \multicolumn{1}{c}{\textbf{Mean}} & \multicolumn{5}{c}{\textbf{Significantly worse results}} & \multicolumn{1}{c}{\textbf{\#}} \\ \hline
\textbf{None} & N & 69.7 & OT & MB & MG &  &  &  & 3 & 50.1 & AT & GB & MB & BH** & MG & 5 \\ \hline
\textbf{Adaptive Thresholding} & AT & 68.5 & MB & MG &  &  &  &  & 2 & 47.2 & MB & MG &  &  &  & 2 \\
\textbf{Otsu Thresholding} & OT & 69.4 & MB & MG &  &  &  &  & 2 & 49.6 & AT & MB & MG &  &  & 3 \\
\textbf{Simple Thresholding} & ST & 69.0 & MB & MG &  &  &  &  & 2 & 49.1 & AT & MB & MG &  &  & 3 \\
\textbf{Triangle Thresholding} & TT & 65.4 & MB & MG &  &  &  &  & 2 & 47.2 & MB & MG &  &  &  & 2 \\ \hline
\textbf{Bilateral Filter} & BF & \textbf{70.7} & AT & ST* & TT** & GB** & MB & BH* & 6 & \textbf{50.3} & AT & GB & MB & BH** & MG & 5 \\
\textbf{Gaussian Blur} & GB & 68.3 & MB & MG &  &  &  &  & 2 & 48.0 & MB & MG &  &  &  & 2 \\
\textbf{Homogeneous Blur} & HB & 69.7 & OT** & MB & MG &  &  &  & 3 & 50.1 & AT & GB & MB & BH** & MG & 5 \\
\textbf{Median Blur} & MB & 30.0 &  &  &  &  &  &  & 0 & 16.7 & MG &  &  &  &  & 1 \\ \hline
\textbf{Black Hat} & BH & 69.5 & MB & MG &  &  &  &  & 2 & 49.0 & AT** & MB & MG &  &  & 3 \\
\textbf{Closing} & C & 69.7 & OT** & MB & MG &  &  &  & 3 & 50.1 & AT & GB & MB & BH** & MG & 5 \\
\textbf{Dilation} & D & 69.7 & OT** & MB & MG &  &  &  & 3 & 50.1 & AT & GB & MB & BH** & MG & 5 \\
\textbf{Erosion} & E & 69.7 & OT** & MB & MG &  &  &  & 3 & 50.1 & AT & GB & MB & BH** & MG & 5 \\
\textbf{Morphological Gradient} & MG & 52.7 & MB &  &  &  &  &  & 1 & 23.3 & MB &  &  &  &  & 1 \\
\textbf{Opening} & O & 69.7 & OT** & MB & MG &  &  &  & 3 & 50.1 & AT & GB & MB & BH** & MG & 5 \\
\textbf{Top Hat} & TH & 69.5 & MB & MG &  &  &  &  & 2 & 49.8 & AT & GB* & MB & MG &  & 4 \\ \hline
\textbf{$\chi^2$} & \multicolumn{1}{l}{} & 746.29 & \multicolumn{1}{l}{} & \multicolumn{1}{l}{} & \multicolumn{1}{l}{} & \multicolumn{1}{l}{} & \multicolumn{1}{l}{} & \multicolumn{1}{l}{} & \multicolumn{1}{l}{} & 888.56 & \multicolumn{1}{l}{} & \multicolumn{1}{l}{} & \multicolumn{1}{l}{} & \multicolumn{1}{l}{} & \multicolumn{1}{l}{} & \multicolumn{1}{l}{} \\
\textbf{p-value} & \multicolumn{1}{l}{} & \textless 2.2e-16 & \multicolumn{1}{l}{} & \multicolumn{1}{l}{} & \multicolumn{1}{l}{} & \multicolumn{1}{l}{} & \multicolumn{1}{l}{} & \multicolumn{1}{l}{} & \multicolumn{1}{l}{} & \textless 2.2e-16 & \multicolumn{1}{l}{} & \multicolumn{1}{l}{} & \multicolumn{1}{l}{} & \multicolumn{1}{l}{} & \multicolumn{1}{l}{} & \multicolumn{1}{l}{} \\ \hline
\end{tabular}%
}
\\
\tiny \raggedright *P<0.05; **P<0.01
\end{table}

We can observe that from the performance results of binarization methods, Adaptive Thresholding and Otsu Thresholding impair the OCR performance.
For the smoothing algorithms' results, we see that Bilateral Filter is the best-performing algorithm. However, it does not improve the OCR performance. On the other hand, Median Blur and Gaussian Blur worsen the OCR performance.
The results for morphological transformation algorithms indicate that Morphological Gradient and Black Hat deteriorate the performance.

\subsubsection{Theatre plays' covers}

This section presents the results regarding the theatre plays' covers typology. Table~\ref{tab:results_theatre_covers_dataset} presents the results of the statistical tests showing a significantly lower mean OCR performance than in the overall dataset. Furthermore, unlike the results in the overall dataset and the previous typologies where binarization algorithms worsened the OCR performance, in this sample it is a binarization algorithm, Adaptive Thresholding, that is the best-performing algorithm. In sum, three algorithms improved character accuracy: Adaptive Thresholding, Black Hat, and Top Hat.

\begin{table}[ht]
\caption{Summary of Friedman test results and Wilcoxon pairwise comparisons for post-hoc analysis of OCR performance of image processing algorithms with the theatre plays' covers sample. The bold values represent the image processing algorithms and average results that improved the OCR performance.}
\label{tab:results_theatre_covers_dataset}
\resizebox{\textwidth}{!}{%
\begin{tabular}{lcclllllllllrclllr}
\hline
\multirow{2}{*}{} & \multicolumn{1}{l}{\multirow{2}{*}{}} & \multicolumn{11}{c}{\textbf{Character Accuracy}} & \multicolumn{5}{c}{\textbf{F1-score}} \\ \cline{3-18} 
 & \multicolumn{1}{l}{} & \multicolumn{1}{c}{\textbf{Mean}} & \multicolumn{9}{c}{\textbf{Significantly worse results}} & \multicolumn{1}{c}{\textbf{\#}} & \multicolumn{1}{c}{\textbf{Mean}} & \multicolumn{3}{c}{\textbf{Significantly worse results}} & \multicolumn{1}{c}{\textbf{\#}} \\ \hline
\textbf{None} & N & 18.2 & MG &  &  &  &  &  &  &  &  & 1 & 16.6 & MG &  &  & 1 \\ \hline
\textbf{Adaptive Thresholding} & AT & \textbf{35.6} & \textbf{N} & OT* & ST* & BF* & HB* & C* & D* & E* & MG & 9 & \textbf{21.8} & HB* & MB* & MG & 3 \\
\textbf{Otsu Thresholding} & OT & \textbf{22.6} & MG &  &  &  &  &  &  &  &  & 1 & 14.1 & MG &  &  & 1 \\
\textbf{Simple Thresholding} & ST & \textbf{21.4} & MG &  &  &  &  &  &  &  &  & 1 & 14.6 & MG &  &  & 1 \\
\textbf{Triangle Thresholding} & TT & \textbf{26.7} & MG &  &  &  &  &  &  &  &  & 1 & 16.0 & MG &  &  & 1 \\ \hline
\textbf{Bilateral Filter} & BF & \textbf{23.4} & MG &  &  &  &  &  &  &  &  & 1 & 16.1 & MG &  &  & 1 \\
\textbf{Gaussian Blur} & GB & \textbf{21.4} & MG &  &  &  &  &  &  &  &  & 1 & 14.0 & MG &  &  & 1 \\
\textbf{Homogeneous Blur} & HB & \textbf{21.2} & MG &  &  &  &  &  &  &  &  & 1 & 12.3 & MG &  &  & 1 \\
\textbf{Median Blur} & MB & \textbf{22.8} & MG &  &  &  &  &  &  &  &  & 1 & 11.8 & MG &  &  & 1 \\ \hline
\textbf{Black Hat} & BH & \textbf{31.8} & \textbf{N*} & MG &  &  &  &  &  &  &  & 2 & \textbf{22.2} & MG &  &  & 1 \\
\textbf{Closing} & C & \textbf{22.4} & MG &  &  &  &  &  &  &  &  & 1 & 14.7 & MG &  &  & 1 \\
\textbf{Dilation} & D & \textbf{22.4} & MG &  &  &  &  &  &  &  &  & 1 & 14.7 & MG &  &  & 1 \\
\textbf{Erosion} & E & \textbf{22.4} & MG &  &  &  &  &  &  &  &  & 1 & 14.7 & MG &  &  & 1 \\
\textbf{Morphological Gradient} & MG & 0.47 &  &  &  &  &  &  &  &  &  & 0 & 0 &  &  &  & 0 \\
\textbf{Opening} & O & \textbf{25.7} & MG &  &  &  &  &  &  &  &  & 1 & \textbf{16.3} & MG &  &  & 1 \\
\textbf{Top Hat} & TH & \textbf{29.3} & \textbf{N*} & MG &  &  &  &  &  &  &  & 2 & \textbf{18.9} & MG &  &  & 1 \\ \hline
\textbf{$\chi^2$} & \multicolumn{1}{l}{} & 92.122 & \multicolumn{1}{l}{} & \multicolumn{1}{l}{} & \multicolumn{1}{l}{} & \multicolumn{1}{l}{} & \multicolumn{1}{l}{} & \multicolumn{1}{l}{} &  &  &  & \multicolumn{1}{l}{} & 81.827 & \multicolumn{1}{l}{} & \multicolumn{1}{l}{} & \multicolumn{1}{l}{} & \multicolumn{1}{l}{} \\
\textbf{p-value} & \multicolumn{1}{l}{} & 1.506e-13 & \multicolumn{1}{l}{} & \multicolumn{1}{l}{} & \multicolumn{1}{l}{} & \multicolumn{1}{l}{} & \multicolumn{1}{l}{} & \multicolumn{1}{l}{} &  &  &  & \multicolumn{1}{l}{} & 1.295e-11 & \multicolumn{1}{l}{} & \multicolumn{1}{l}{} & \multicolumn{1}{l}{} & \multicolumn{1}{l}{} \\ \hline
\end{tabular}%
}
\\
\tiny \raggedright *P<0.05; **P<0.01
\end{table}

\subsection{OCR Performance Comparison Across Typologies and Overall} \label{perf_comparison}

\begin{figure}[p]
 \rotatebox{90}{%
     \begin{minipage}{\textheight}%
    \centering
    \begin{subfigure}{\linewidth}
        \includegraphics[width=\linewidth]{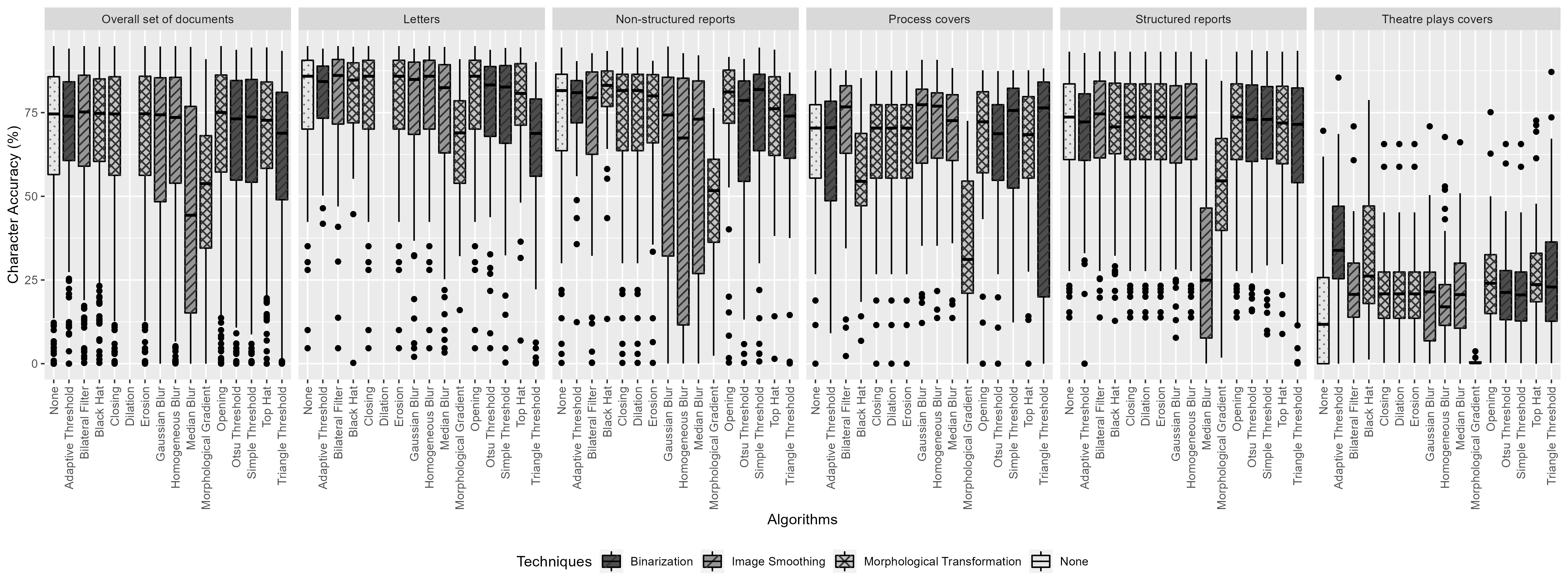}
    \end{subfigure}
    \begin{subfigure}{\linewidth}
        \includegraphics[width=\linewidth]{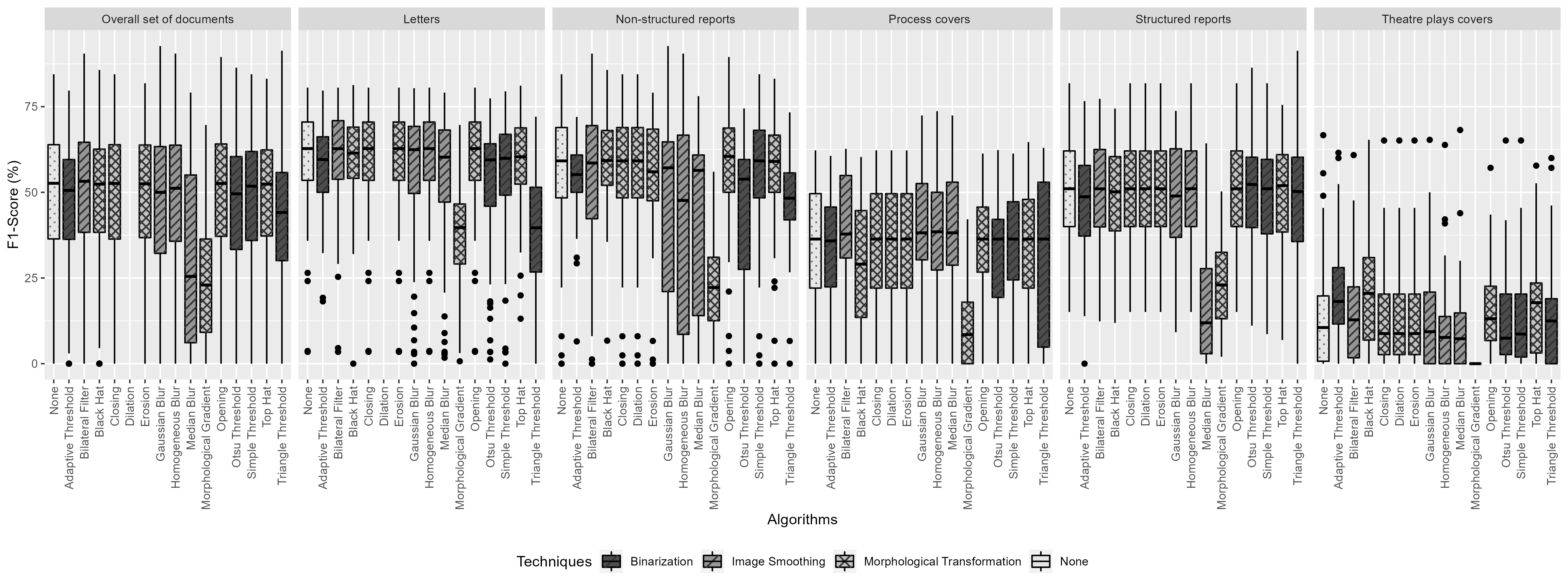}
    \end{subfigure}
    \caption{Distribution of OCR performance measures in the overall set of documents, and by letters, non-structured reports, process covers, structured reports, and theatre plays' covers. The first row of results refers to the character accuracy performance measure, and the second row refers to the F1-score performance measure.}
    \label{fig:distribution}
    \end{minipage}%
  }%
\end{figure}

We present the distribution of the performance measures for the overall dataset and the digital representation-type samples in Figure~\ref{fig:distribution}. We conclude that the digital representation typology with the best results is the letters sample, with a median character accuracy above 75\% and a median F1-score above 60\% for most algorithms. In contrast, the digital representation typology with the worst performance results is the theatre plays' covers sample, with a median character accuracy below 25\% and a median F1-score below 20\%. Image processing algorithms appear to have a greater impact on the OCR performance of digital representations from the theatre plays' covers sample, where the OCR performance without applying image processing algorithms is low.

\section{Discussion} \label{sec:discussion}

Our results from Sections~\ref{best_parameters} and~\ref{default_effects} show that parameterization with the NSGA-II algorithm influences the impact of image processing algorithms on OCR, in response to RQ1. Not only are the parameter values very different with parameterization, but the results are significantly better compared to using default parameter values. We conclude that some image processing algorithms perform significantly better with parameterization by typology than with global parameterization. Moreover, parameterization by typology has the advantage of being faster to process than global parameterization.

From Section~\ref{compare_overall}, we can draw conclusions to answer RQ2. We conclude that there are no significant improvements in the OCR performance in the overall dataset when using binarization algorithms in an image pre-processing OCR phase in comparison to not using them. Unlike the binarization method, the smoothing and morphological transformation methods significantly improved the OCR performance with Bilateral Filter and Opening, respectively. Table~\ref{tab:algorithm_selection} shows the image processing algorithms we chose to apply as image pre-processing to OCR from the results we described in Sections~\ref{perf_analysis} and~\ref{compare_type}. In the column ``Improved algorithms'' of Table~\ref{tab:algorithm_selection}, we list the image processing algorithms that significantly improved the OCR performance without image pre-processing. From these algorithms, we selected the ones that had the higher significance levels for the overall dataset and each digital representation typology. The two last columns of Table~\ref{tab:algorithm_selection} show the mean performance values of the selected algorithms. The OCR type error analysis in Section~\ref{error_analysis} lead us to conclude that among single modification error types, the operation of deletion is the most common, and insertion is the least common. Moreover, among the combination of edit operation types, the operations of deletion and insertion, along with deletion, insertion, and substitution have a very small occurrence. We observe that image processing algorithms that performed the best in the overall dataset had a lower occurrence of deletion edit operations, and those that performed the worst had the highest occurrence. This might signify that 1) applying image pre-processing to OCR has a bigger impact in mitigating the error of missing text in comparison to other errors, and 2) improving the lack of text recognition has a bigger positive influence in improving the OCR performance than improving other errors.

\begin{table}[ht]
\caption{Image processing algorithms that improved the OCR performance without image pre-processing per digital representation typology and overall and were selected as the most significant. The mean OCR performance values refer to the selected image pre-processing algorithm.}
\label{tab:algorithm_selection}
\begin{tabular}{lllcc}
\hline
\textbf{Typology} & \textbf{Improved algorithms} & \textbf{Selected algorithm} & \textbf{\begin{tabular}[c]{@{}c@{}}Mean character\\ accuracy\end{tabular}} & \textbf{\begin{tabular}[c]{@{}c@{}}Mean\\ F1-score\end{tabular}} \\ \hline
Overall & Bilateral Filter and Opening & Opening & 67.8\% & 49.2\% \\ \hline
Letters & Bilateral Filter & Bilateral Filter & 78.6\% & 60.2\% \\ \hline
\begin{tabular}[c]{@{}l@{}}Non-structured reports\end{tabular} & None & None & 69.2\% & 53.8\% \\ \hline
Process covers & None & None & 62.4\% & 35.3\% \\ \hline
Structured reports & None & None & 69.7\% & 50.1\% \\ \hline
\begin{tabular}[c]{@{}l@{}}Theatre plays' covers\end{tabular} & \begin{tabular}[c]{@{}l@{}}Adaptive Threshold, Black Hat\\ and Top Hat\end{tabular} & Adaptive Threshold & 35.6\% & 21.8\% \\ \hline
\end{tabular}
\end{table}

Results from Section~\ref{compare_type} allow us to answer RQ3. The analysis shows that it is possible to improve the text recognition task using tuned image processing algorithms in the subsets of letters and theatre plays' covers. As shown in the column ``Improved algorithms'' of Table~\ref{tab:algorithm_selection}, Bilateral Filter significantly improved the OCR performance with image pre-processing with the subset of letters, and Adaptive Threshold, Black Hat, and Top Hat improved the performance with the subset of theatre plays' covers. However, no image processing algorithms significantly improved the OCR performance with the samples of all the other typologies.
From the column ``Selected algorithm'' of Table~\ref{tab:algorithm_selection}, we can see that the algorithm we chose for the typology of letters is Bilateral Filter, and for theatre plays' covers it is Adaptive Threshold.
From the performance comparison in Section~\ref{perf_comparison}, we conclude that process covers and theatre plays' covers are the samples with the lowest mean OCR performance values. They are also the samples with the most algorithms that tend to improve the OCR performance or do so in a significant way. These findings suggest that image processing algorithms tend to have a greater impact on digital representation typologies that have a lower OCR performance without an image pre-processing phase. The results also show that morphological transformation algorithms tend to be the most beneficial to every typology. In contrast, binarization algorithms only benefit the theatre plays' covers typology and tend to do so for the sample of process covers. On the other hand, while smoothing algorithms tend to be beneficial to every typology, like morphological transformation algorithms, only one algorithm, Bilateral Filter, is responsible for that performance improvement.

\section{Conclusion} \label{conclusion}

The main goal of this work is to understand how to improve the OCR performance on cultural heritage digital representations using image processing algorithms. We evaluated the use and parameterization of image processing algorithms for text recognition tasks using Tesseract and OpenCV methods.

The key takeaway is that parameterization by digital representation typology benefits the employment of image pre-processing techniques in OCR, even if only for a small parameterization dataset. We successfully applied the NSGA-II algorithm to perform the parameterization. The analysis of the performance of image processing algorithms leads us to conclude that some algorithms improve the OCR performance significantly, while several tend to improve it and it is up to debate whether to apply them or not, and others even worsen the performance. A previous analysis is required before applying them. On top of that, some algorithms have more impact in some samples of the evaluation dataset than others. Digital representation typologies influence the effect these image processing algorithms have on the OCR performance. For this reason, it is important to analyze typologies before applying OCR. A binarization algorithm positively impacted the OCR performance in the sample of theatre plays' covers. On the other hand, smoothing algorithms improved the performance of the overall dataset and the samples of letters and theatre plays' covers. Likewise, morphological transformation algorithms also improved the OCR performance of the overall dataset and the samples of theatre plays' covers. These findings suggest that the employment of image pre-processing algorithms in OCR might be more suitable for typologies where the text recognition task without pre-processing does not produce good results, as is the case of the typology for theatre plays' covers. Factually, the algorithms that have been proven to improve the OCR task compared to not applying any are Adaptive Thresholding, Bilateral Filter, Black Hat, Opening, and Top Hat.  Ultimately, there are big differences in the OCR performance by digital representation typology, which suggest that we should adopt the optimization of the OCR task for typologies that have weaker results without employing image pre-processing algorithms.

The practical outcomes of this work are the source code of the parameterization of image processing algorithms and two datasets. Both datasets are available to consult for other scholars or researchers to use. Likewise, we provide the source code of the parameterization of image processing algorithms using NSGA-II.

In the future, we will analyze the effect of the combination of image processing algorithms in the text recognition task. We would also like to expand the current OCR error analysis to include other error types. Furthermore, we would like to explore the detection and removal of stamps and annotations in digital representations in the OCR performance.

\begin{acks}
This work is financed by National Funds through FCT - Foundation for Science and Technology I.P., within the scope of the EPISA project - DSAIPA/DS/0023/2018.
\end{acks}

\bibliographystyle{ACM-Reference-Format}
\bibliography{sample-base}


\begin{thebibliography}{47}


\ifx \showCODEN    \undefined \def \showCODEN     #1{\unskip}     \fi
\ifx \showDOI      \undefined \def \showDOI       #1{#1}\fi
\ifx \showISBNx    \undefined \def \showISBNx     #1{\unskip}     \fi
\ifx \showISBNxiii \undefined \def \showISBNxiii  #1{\unskip}     \fi
\ifx \showISSN     \undefined \def \showISSN      #1{\unskip}     \fi
\ifx \showLCCN     \undefined \def \showLCCN      #1{\unskip}     \fi
\ifx \shownote     \undefined \def \shownote      #1{#1}          \fi
\ifx \showarticletitle \undefined \def \showarticletitle #1{#1}   \fi
\ifx \showURL      \undefined \def \showURL       {\relax}        \fi
\providecommand\bibfield[2]{#2}
\providecommand\bibinfo[2]{#2}
\providecommand\natexlab[1]{#1}
\providecommand\showeprint[2][]{arXiv:#2}

\bibitem[Antonacopoulos et~al\mbox{.}(2013)]%
        {antonacopoulos2013}
\bibfield{author}{\bibinfo{person}{Apostolos Antonacopoulos},
  \bibinfo{person}{Christian Clausner}, \bibinfo{person}{Christos
  Papadopoulos}, {and} \bibinfo{person}{Stefan Pletschacher}.}
  \bibinfo{year}{2013}\natexlab{}.
\newblock \showarticletitle{{ICDAR 2013 Competition on Historical Book
  Recognition (HBR 2013)}}. In \bibinfo{booktitle}{\emph{Proceedings of the
  2013 12th International Conference on Document Analysis and Recognition}}
  \emph{(\bibinfo{series}{ICDAR '13})}. \bibinfo{publisher}{IEEE Computer
  Society}, \bibinfo{address}{USA}, \bibinfo{pages}{1459–1463}.
\newblock
\showISBNx{9780769549996}
\urldef\tempurl%
\url{https://doi.org/10.1109/ICDAR.2013.294}
\showDOI{\tempurl}


\bibitem[Baierer(2020)]%
        {ocropus}
\bibfield{author}{\bibinfo{person}{Konstantin Baierer}.}
  \bibinfo{year}{2020}\natexlab{}.
\newblock \bibinfo{title}{{Models · OCROPUS/ocropy wiki}}.
\newblock
\newblock
\urldef\tempurl%
\url{https://github.com/ocropus/ocropy/wiki/Models}
\showURL{%
\tempurl}


\bibitem[Bizer et~al\mbox{.}(2009)]%
        {bizer2009}
\bibfield{author}{\bibinfo{person}{Christian Bizer}, \bibinfo{person}{Tom
  Heath}, {and} \bibinfo{person}{Tim Berners-Lee}.}
  \bibinfo{year}{2009}\natexlab{}.
\newblock \showarticletitle{{Linked Data: The Story so Far}}.
\newblock \bibinfo{journal}{\emph{{International Journal on Semantic Web and
  Information Systems}}}  \bibinfo{volume}{5} (\bibinfo{date}{07}
  \bibinfo{year}{2009}), \bibinfo{pages}{1--22}.
\newblock
\urldef\tempurl%
\url{https://doi.org/10.4018/jswis.2009081901}
\showDOI{\tempurl}


\bibitem[Blank and Deb(2020)]%
        {blank2020}
\bibfield{author}{\bibinfo{person}{Julian Blank} {and} \bibinfo{person}{Kalyan
  Deb}.} \bibinfo{year}{2020}\natexlab{}.
\newblock \showarticletitle{{Pymoo: Multi-Objective Optimization in Python}}.
\newblock \bibinfo{journal}{\emph{{IEEE Access}}}  \bibinfo{volume}{PP}
  (\bibinfo{date}{04} \bibinfo{year}{2020}), \bibinfo{pages}{1--1}.
\newblock
\urldef\tempurl%
\url{https://doi.org/10.1109/ACCESS.2020.2990567}
\showDOI{\tempurl}


\bibitem[Bromage(2016)]%
        {bromage2016}
\bibfield{author}{\bibinfo{person}{David Bromage}.}
  \bibinfo{year}{2016}\natexlab{}.
\newblock \showarticletitle{{Linked Data for Libraries, Archives and Museums:
  How to Clean, Link and Publish Your Metadata}}.
\newblock \bibinfo{journal}{\emph{Archives and Manuscripts}}
  \bibinfo{volume}{44}, \bibinfo{number}{3} (\bibinfo{date}{Oct.}
  \bibinfo{year}{2016}), \bibinfo{pages}{172--174}.
\newblock
\urldef\tempurl%
\url{https://doi.org/10.1080/01576895.2016.1233606}
\showDOI{\tempurl}


\bibitem[Bui et~al\mbox{.}(2017)]%
        {bui2017}
\bibfield{author}{\bibinfo{person}{Quang~Anh Bui}, \bibinfo{person}{David
  Mollard}, {and} \bibinfo{person}{Salvatore Tabbone}.}
  \bibinfo{year}{2017}\natexlab{}.
\newblock \showarticletitle{{Selecting Automatically Pre-Processing Methods to
  Improve OCR Performances}}. In \bibinfo{booktitle}{\emph{{2017 14th IAPR
  International Conference on Document Analysis and Recognition (ICDAR)}}},
  Vol.~\bibinfo{volume}{01}. \bibinfo{publisher}{IEEE Computer Society},
  \bibinfo{address}{Kyoto, Japan}, \bibinfo{pages}{169--174}.
\newblock
\urldef\tempurl%
\url{https://doi.org/10.1109/ICDAR.2017.36}
\showDOI{\tempurl}


\bibitem[Carrasco(2014)]%
        {Carrasco2014}
\bibfield{author}{\bibinfo{person}{Rafael~C. Carrasco}.}
  \bibinfo{year}{2014}\natexlab{}.
\newblock \showarticletitle{{An Open-Source OCR Evaluation Tool}}. In
  \bibinfo{booktitle}{\emph{Proceedings of the First International Conference
  on Digital Access to Textual Cultural Heritage}} (Madrid, Spain)
  \emph{(\bibinfo{series}{DATeCH '14})}. \bibinfo{publisher}{Association for
  Computing Machinery}, \bibinfo{address}{New York, NY, USA},
  \bibinfo{pages}{179–184}.
\newblock
\showISBNx{9781450325882}
\urldef\tempurl%
\url{https://doi.org/10.1145/2595188.2595221}
\showDOI{\tempurl}


\bibitem[Clausner et~al\mbox{.}(2013)]%
        {clausner2013}
\bibfield{author}{\bibinfo{person}{Christian Clausner}, \bibinfo{person}{Stefan
  Pletschacher}, {and} \bibinfo{person}{Apostolos Antonacopoulos}.}
  \bibinfo{year}{2013}\natexlab{}.
\newblock \showarticletitle{The Significance of Reading Order in Document
  Recognition and Its Evaluation}. In \bibinfo{booktitle}{\emph{2013 12th
  International Conference on Document Analysis and Recognition}}.
  \bibinfo{publisher}{{IEEE Computer Society}}, \bibinfo{address}{Washington,
  DC, USA}, \bibinfo{pages}{688--692}.
\newblock
\urldef\tempurl%
\url{https://doi.org/10.1109/ICDAR.2013.141}
\showDOI{\tempurl}


\bibitem[Clausner et~al\mbox{.}(2020)]%
        {clausner2020}
\bibfield{author}{\bibinfo{person}{Christian Clausner}, \bibinfo{person}{Stefan
  Pletschacher}, {and} \bibinfo{person}{Apostolos Antonacopoulos}.}
  \bibinfo{year}{2020}\natexlab{}.
\newblock \showarticletitle{{Flexible character accuracy measure for
  reading-order-independent evaluation}}.
\newblock \bibinfo{journal}{\emph{Pattern Recognition Letters}}
  \bibinfo{volume}{131} (\bibinfo{year}{2020}), \bibinfo{pages}{390--397}.
\newblock
\showISSN{0167-8655}
\urldef\tempurl%
\url{https://doi.org/10.1016/j.patrec.2020.02.003}
\showDOI{\tempurl}


\bibitem[DGARQ(2008)]%
        {digitarq}
\bibfield{author}{\bibinfo{person}{DGARQ}.} \bibinfo{year}{2008}\natexlab{}.
\newblock \bibinfo{title}{{Arquivo Nacional Torre do Tombo}}.
\newblock
\newblock
\urldef\tempurl%
\url{https://digitarq.arquivos.pt/}
\showURL{%
\tempurl}


\bibitem[Dias(2022)]%
        {dias2022}
\bibfield{author}{\bibinfo{person}{Mariana Dias}.}
  \bibinfo{year}{2022}\natexlab{}.
\newblock \bibinfo{title}{{Typewritten Digital Representations of Portuguese
  Cultural Heritage Documents from the 20th century}}.
\newblock \bibinfo{howpublished}{Data set}.
\newblock
\urldef\tempurl%
\url{https://doi.org/10.25747/ZC25-1531}
\showDOI{\tempurl}


\bibitem[Dias and Lopes(2022)]%
        {dias_article2022}
\bibfield{author}{\bibinfo{person}{Mariana Dias} {and}
  \bibinfo{person}{Carla~Teixeira Lopes}.} \bibinfo{year}{2022}\natexlab{}.
\newblock \showarticletitle{{Mining Typewritten Digital Representations to
  Support Archival Description}}. In \bibinfo{booktitle}{\emph{Proceedings of
  the 26th International Conference on Theory and Practice of Digital Libraries
  - Workshops and Doctoral Consortium}} \emph{(\bibinfo{series}{{CEUR} Workshop
  Proceedings}, Vol.~\bibinfo{volume}{3246})},
  \bibfield{editor}{\bibinfo{person}{Leonardo Candela} {and}
  \bibinfo{person}{Gianmaria Silvello}} (Eds.).
  \bibinfo{publisher}{CEUR-WS.org}, \bibinfo{address}{Padua, Italy},
  \bibinfo{pages}{70--76}.
\newblock
\urldef\tempurl%
\url{https://ceur-ws.org/Vol-3246/09\_Paper2.pdf}
\showURL{%
\tempurl}


\bibitem[dos Arquivos Nacionais/Torre~do Tombo(2006)]%
        {dgarq2006}
\bibfield{author}{\bibinfo{person}{Instituto dos Arquivos Nacionais/Torre~do
  Tombo}.} \bibinfo{year}{2006}\natexlab{}.
\newblock \bibinfo{title}{{Orientações para a gestão de documentos de
  Arquivo no contexto de uma reestruturação da Administração Central do
  Estado}}.
\newblock , \bibinfo{numpages}{55}~pages.
\newblock


\bibitem[Drobac and Lindén(2020)]%
        {drobac2020}
\bibfield{author}{\bibinfo{person}{Senka Drobac} {and} \bibinfo{person}{Krister
  Lindén}.} \bibinfo{year}{2020}\natexlab{}.
\newblock \showarticletitle{{Optical character recognition with neural networks
  and post-correction with finite state methods}}.
\newblock \bibinfo{journal}{\emph{{International Journal on Document Analysis
  and Recognition (IJDAR)}}}  \bibinfo{volume}{23} (\bibinfo{date}{12}
  \bibinfo{year}{2020}), \bibinfo{pages}{1--17}.
\newblock
\urldef\tempurl%
\url{https://doi.org/10.1007/s10032-020-00359-9}
\showDOI{\tempurl}


\bibitem[Falcão(2022)]%
        {falcao2022}
\bibfield{author}{\bibinfo{person}{Margarida Falcão}.}
  \bibinfo{year}{2022}\natexlab{}.
\newblock \bibinfo{title}{{Manual Transcriptions of Typewritten Digital
  Representations of Portuguese Cultural Heritage Documents from the 20th
  Century}}.
\newblock \bibinfo{howpublished}{Data set}.
\newblock
\urldef\tempurl%
\url{https://doi.org/10.25747/WPNA-JE39}
\showDOI{\tempurl}


\bibitem[{FineReader PDF}(2019)]%
        {finereader2019}
\bibfield{author}{\bibinfo{person}{{FineReader PDF}}.}
  \bibinfo{year}{2019}\natexlab{}.
\newblock \bibinfo{title}{{How AI powers PDF Software \& Technology Trends:
  FineReader Blog}}.
\newblock
\newblock
\urldef\tempurl%
\url{https://pdf.abbyy.com/blog/finereader-powered-by-ai/}
\showURL{%
\tempurl}


\bibitem[{FineReader PDF}(2021)]%
        {finereader2021}
\bibfield{author}{\bibinfo{person}{{FineReader PDF}}.}
  \bibinfo{year}{2021}\natexlab{}.
\newblock \bibinfo{title}{{Technical Specifications and System Requirements |
  FineReader PDF}}.
\newblock
\newblock
\urldef\tempurl%
\url{https://pdf.abbyy.com/specifications/}
\showURL{%
\tempurl}


\bibitem[Fontan et~al\mbox{.}(2016)]%
        {fontan2016}
\bibfield{author}{\bibinfo{person}{Lionel Fontan}, \bibinfo{person}{Isabelle
  Ferran{\'e}}, \bibinfo{person}{J{\'e}r{\^o}me Farinas},
  \bibinfo{person}{Julien Pinquier}, {and} \bibinfo{person}{Xavier Aumont}.}
  \bibinfo{year}{2016}\natexlab{}.
\newblock \showarticletitle{{Using Phonologically Weighted Levenshtein
  Distances for the Prediction of Microscopic Intelligibility}}. In
  \bibinfo{booktitle}{\emph{{Interspeech 2016, 17th Annual Conference of the
  International Speech Communication Association}}}.
  \bibinfo{publisher}{{ISCA}}, \bibinfo{address}{San Francisco, CA, United
  States}, \bibinfo{pages}{650--654}.
\newblock
\urldef\tempurl%
\url{https://doi.org/10.21437/Interspeech.2016-431}
\showDOI{\tempurl}


\bibitem[Gao et~al\mbox{.}(2019)]%
        {gao2019}
\bibfield{author}{\bibinfo{person}{Yayu Gao}, \bibinfo{person}{Xinmin Zhang},
  \bibinfo{person}{Xiaoyou Zhang}, \bibinfo{person}{Duan Li},
  \bibinfo{person}{Min Yang}, {and} \bibinfo{person}{Jinhua Tian}.}
  \bibinfo{year}{2019}\natexlab{}.
\newblock \showarticletitle{{Application of NSGA-II and Improved Risk Decision
  Method for Integrated Water Resources Management of Malian River Basin}}.
\newblock \bibinfo{journal}{\emph{Water}} \bibinfo{volume}{11},
  \bibinfo{number}{8} (\bibinfo{year}{2019}), \bibinfo{numpages}{27}~pages.
\newblock
\showISSN{2073-4441}
\urldef\tempurl%
\url{https://doi.org/10.3390/w11081650}
\showDOI{\tempurl}


\bibitem[Gupta et~al\mbox{.}(2007)]%
        {gupta2007}
\bibfield{author}{\bibinfo{person}{Maya Gupta}, \bibinfo{person}{Nathaniel
  Jacobson}, {and} \bibinfo{person}{Eric Garcia}.}
  \bibinfo{year}{2007}\natexlab{}.
\newblock \showarticletitle{{OCR binarization and image pre-processing for
  searching historical documents}}.
\newblock \bibinfo{journal}{\emph{{Pattern Recognition}}}  \bibinfo{volume}{40}
  (\bibinfo{date}{02} \bibinfo{year}{2007}), \bibinfo{pages}{389--397}.
\newblock
\urldef\tempurl%
\url{https://doi.org/10.1016/j.patcog.2006.04.043}
\showDOI{\tempurl}


\bibitem[Harraj and Raissouni(2015)]%
        {harraj2015}
\bibfield{author}{\bibinfo{person}{Abdeslam Harraj} {and}
  \bibinfo{person}{Naoufal Raissouni}.} \bibinfo{year}{2015}\natexlab{}.
\newblock \showarticletitle{{OCR Accuracy Improvement on Document Images
  Through a Novel Pre-Processing Approach}}.
\newblock \bibinfo{journal}{\emph{{Signal \& Image Processing : An
  International Journal}}}  \bibinfo{volume}{6} (\bibinfo{date}{09}
  \bibinfo{year}{2015}).
\newblock
\urldef\tempurl%
\url{https://doi.org/10.5121/sipij.2015.6401}
\showDOI{\tempurl}


\bibitem[Kiss et~al\mbox{.}(2019)]%
        {kiss2019}
\bibfield{author}{\bibinfo{person}{Martin Kiss}, \bibinfo{person}{Michal
  Hradis}, {and} \bibinfo{person}{Oldrich Kodym}.}
  \bibinfo{year}{2019}\natexlab{}.
\newblock \showarticletitle{{Brno Mobile {OCR} Dataset}}.
\newblock \bibinfo{journal}{\emph{{CoRR}}}  \bibinfo{volume}{abs/1907.01307}
  (\bibinfo{year}{2019}), \bibinfo{numpages}{6}~pages.
\newblock
\urldef\tempurl%
\url{http://arxiv.org/abs/1907.01307}
\showURL{%
\tempurl}


\bibitem[Koistinen et~al\mbox{.}(2017)]%
        {koistinen2017}
\bibfield{author}{\bibinfo{person}{Mika Koistinen}, \bibinfo{person}{Kimmo
  Kettunen}, {and} \bibinfo{person}{Tuula Pääkkönen}.}
  \bibinfo{year}{2017}\natexlab{}.
\newblock \showarticletitle{{Improving Optical Character Recognition of Finnish
  Historical Newspapers with a Combination of Fraktur \& Antiqua Models and
  Image Preprocessing}}. In \bibinfo{booktitle}{\emph{{Proceedings of the 21st
  Nordic Conference on Computational Linguistics}}}.
  \bibinfo{publisher}{{Association for Computational Linguistics}},
  \bibinfo{address}{Gothenburg, Sweden}, \bibinfo{pages}{277--283}.
\newblock
\urldef\tempurl%
\url{https://aclanthology.org/W17-0238}
\showURL{%
\tempurl}


\bibitem[Kotthoff(2016)]%
        {kotthoff2016}
\bibfield{author}{\bibinfo{person}{Lars Kotthoff}.}
  \bibinfo{year}{2016}\natexlab{}.
\newblock \bibinfo{booktitle}{\emph{{Algorithm Selection for Combinatorial
  Search Problems: A Survey}}}.
\newblock \bibinfo{publisher}{{Springer International Publishing}},
  \bibinfo{address}{Cham}, \bibinfo{pages}{149--190}.
\newblock
\urldef\tempurl%
\url{https://doi.org/10.1007/978-3-319-50137-6_7}
\showDOI{\tempurl}


\bibitem[Kumar et~al\mbox{.}(2013)]%
        {kumar2013}
\bibfield{author}{\bibinfo{person}{Gaurav Kumar}, \bibinfo{person}{Pradeep
  Bhatia}, {and} \bibinfo{person}{Indu}.} \bibinfo{year}{2013}\natexlab{}.
\newblock \showarticletitle{{Analytical Review of Preprocessing Techniques for
  Offline Handwritten Character Recognition}}.
\newblock \bibinfo{journal}{\emph{{International Journal of Advances in
  Engineering Sciences}}}  \bibinfo{volume}{3} (\bibinfo{date}{07}
  \bibinfo{year}{2013}), \bibinfo{pages}{14--22}.
\newblock
\urldef\tempurl%
\url{https://doi.org/10.13140/RG.2.1.3896.7842}
\showDOI{\tempurl}


\bibitem[Lund et~al\mbox{.}(2013)]%
        {lund2013}
\bibfield{author}{\bibinfo{person}{William Lund}, \bibinfo{person}{Douglas
  Kennard}, {and} \bibinfo{person}{Eric Ringger}.}
  \bibinfo{year}{2013}\natexlab{}.
\newblock \showarticletitle{{Combining Multiple Thresholding Binarization
  Values to Improve OCR Output}}. In \bibinfo{booktitle}{\emph{{Document
  Recognition and Retrieval XX}}}, Vol.~\bibinfo{volume}{8658}.
  \bibinfo{publisher}{{SPIE}}, \bibinfo{address}{Burlingame, California, USA}.
\newblock
\urldef\tempurl%
\url{https://doi.org/10.1117/12.2006228}
\showDOI{\tempurl}


\bibitem[Lyu et~al\mbox{.}(2021)]%
        {Lyu2021}
\bibfield{author}{\bibinfo{person}{Lijun Lyu}, \bibinfo{person}{Maria
  Koutraki}, \bibinfo{person}{Martin Krickl}, {and} \bibinfo{person}{Besnik
  Fetahu}.} \bibinfo{year}{2021}\natexlab{}.
\newblock \showarticletitle{{Neural {OCR} Post-Hoc Correction of Historical
  Corpora}}.
\newblock \bibinfo{journal}{\emph{CoRR}}  \bibinfo{volume}{abs/2102.00583}
  (\bibinfo{year}{2021}), \bibinfo{numpages}{15}~pages.
\newblock
\urldef\tempurl%
\url{https://arxiv.org/abs/2102.00583}
\showURL{%
\tempurl}


\bibitem[Mohamad(2015)]%
        {mohamad2015}
\bibfield{author}{\bibinfo{person}{Mumtazimah Mohamad}.}
  \bibinfo{year}{2015}\natexlab{}.
\newblock \bibinfo{title}{{A Review on OpenCV}}.
\newblock
\newblock
\urldef\tempurl%
\url{https://doi.org/10.13140/RG.2.1.2269.8721}
\showDOI{\tempurl}


\bibitem[Nebro et~al\mbox{.}(2022)]%
        {nebro2022}
\bibfield{author}{\bibinfo{person}{Antonio~J. Nebro}, \bibinfo{person}{Jesús
  Galeano-Brajones}, \bibinfo{person}{Francisco Luna}, {and}
  \bibinfo{person}{Carlos~A. Coello~Coello}.} \bibinfo{year}{2022}\natexlab{}.
\newblock \showarticletitle{{Is NSGA-II Ready for Large-Scale Multi-Objective
  Optimization?}}
\newblock \bibinfo{journal}{\emph{Mathematical and Computational Applications}}
  \bibinfo{volume}{27}, \bibinfo{number}{6} (\bibinfo{year}{2022}),
  \bibinfo{numpages}{17}~pages.
\newblock
\showISSN{2297-8747}
\urldef\tempurl%
\url{https://doi.org/10.3390/mca27060103}
\showDOI{\tempurl}


\bibitem[Nguyen et~al\mbox{.}(2020)]%
        {Nguyen2019}
\bibfield{author}{\bibinfo{person}{Thi-Tuyet-Hai Nguyen}, \bibinfo{person}{Adam
  Jatowt}, \bibinfo{person}{Mickael Coustaty}, \bibinfo{person}{Nhu-Van
  Nguyen}, {and} \bibinfo{person}{Antoine Doucet}.}
  \bibinfo{year}{2020}\natexlab{}.
\newblock \showarticletitle{{Deep Statistical Analysis of OCR Errors for
  Effective Post-OCR Processing}}. In \bibinfo{booktitle}{\emph{2019 ACM/IEEE
  Joint Conference on Digital Libraries (JCDL)}} \emph{(\bibinfo{series}{JCDL
  '19})}. \bibinfo{publisher}{IEEE Press}, \bibinfo{address}{Champaign,
  Illinois, USA}, \bibinfo{pages}{29–38}.
\newblock
\showISBNx{9781728115474}
\urldef\tempurl%
\url{https://doi.org/10.1109/JCDL.2019.00015}
\showDOI{\tempurl}


\bibitem[Niu(2016)]%
        {niu2016}
\bibfield{author}{\bibinfo{person}{Jinfang Niu}.}
  \bibinfo{year}{2016}\natexlab{}.
\newblock \showarticletitle{{Linked Data for Archives}}.
\newblock \bibinfo{journal}{\emph{Archivaria}}  \bibinfo{volume}{82}
  (\bibinfo{date}{12} \bibinfo{year}{2016}), \bibinfo{pages}{83--110}.
\newblock
\urldef\tempurl%
\url{https://archivaria.ca/index.php/archivaria/article/view/13582}
\showURL{%
\tempurl}


\bibitem[Noether(1992)]%
        {noether1992}
\bibfield{author}{\bibinfo{person}{Gottfried~E. Noether}.}
  \bibinfo{year}{1992}\natexlab{}.
\newblock \bibinfo{booktitle}{\emph{Introduction to Wilcoxon (1945) Individual
  Comparisons by Ranking Methods}}.
\newblock \bibinfo{publisher}{Springer New York}, \bibinfo{address}{New York,
  NY}, \bibinfo{pages}{191--195}.
\newblock
\urldef\tempurl%
\url{https://doi.org/10.1007/978-1-4612-4380-9_15}
\showDOI{\tempurl}


\bibitem[Ntogas and Ventzas(2009)]%
        {ntogas2009}
\bibfield{author}{\bibinfo{person}{Nikolaos Ntogas} {and}
  \bibinfo{person}{Dimitrios Ventzas}.} \bibinfo{year}{2009}\natexlab{}.
\newblock \showarticletitle{{Binarization of pre-filtered historical
  manuscripts images}}.
\newblock \bibinfo{journal}{\emph{{International Journal of Intelligent
  Computing and Cybernetics}}}  \bibinfo{volume}{2} (\bibinfo{date}{03}
  \bibinfo{year}{2009}), \bibinfo{pages}{148--174}.
\newblock
\urldef\tempurl%
\url{https://doi.org/10.1108/17563780910939282}
\showDOI{\tempurl}


\bibitem[Pfaffe et~al\mbox{.}(2017)]%
        {pfaffe2017}
\bibfield{author}{\bibinfo{person}{Philip Pfaffe}, \bibinfo{person}{Martin
  Tillmann}, \bibinfo{person}{Sigmar Walter}, {and} \bibinfo{person}{Walter
  Tichy}.} \bibinfo{year}{2017}\natexlab{}.
\newblock \showarticletitle{{Online-Autotuning in the Presence of Algorithmic
  Choice}}. In \bibinfo{booktitle}{\emph{{2017 IEEE International Parallel and
  Distributed Processing Symposium Workshops (IPDPSW)}}}.
  \bibinfo{publisher}{IEEE Computer Society}, \bibinfo{address}{Lake Buena
  Vista, FL}, \bibinfo{pages}{1379--1388}.
\newblock
\urldef\tempurl%
\url{https://doi.org/10.1109/IPDPSW.2017.28}
\showDOI{\tempurl}


\bibitem[Pletschacher et~al\mbox{.}(2015)]%
        {pletschacher2015}
\bibfield{author}{\bibinfo{person}{Stefan Pletschacher},
  \bibinfo{person}{Christian Clausner}, {and} \bibinfo{person}{Apostolos
  Antonacopoulos}.} \bibinfo{year}{2015}\natexlab{}.
\newblock \showarticletitle{{Europeana Newspapers OCR Workflow Evaluation}}. In
  \bibinfo{booktitle}{\emph{{Proceedings of the 3rd International Workshop on
  Historical Document Imaging and Processing}}} \emph{(\bibinfo{series}{HIP
  '15})}. \bibinfo{publisher}{Association for Computing Machinery},
  \bibinfo{address}{New York, NY, USA}, \bibinfo{pages}{39--46}.
\newblock
\urldef\tempurl%
\url{https://doi.org/10.1145/2809544.2809554}
\showDOI{\tempurl}


\bibitem[Probst et~al\mbox{.}(2019)]%
        {probst2018}
\bibfield{author}{\bibinfo{person}{Philipp Probst}, \bibinfo{person}{Anne-Laure
  Boulesteix}, {and} \bibinfo{person}{Bernd Bischl}.}
  \bibinfo{year}{2019}\natexlab{}.
\newblock \showarticletitle{{Tunability: Importance of Hyperparameters of
  Machine Learning Algorithms}}.
\newblock \bibinfo{journal}{\emph{J. Mach. Learn. Res.}} \bibinfo{volume}{20},
  \bibinfo{number}{1} (\bibinfo{date}{jan} \bibinfo{year}{2019}),
  \bibinfo{pages}{1934–1965}.
\newblock
\showISSN{1532-4435}


\bibitem[Roetzel et~al\mbox{.}(2020)]%
        {roetzel2020}
\bibfield{author}{\bibinfo{person}{Wilfried Roetzel}, \bibinfo{person}{Xing
  Luo}, {and} \bibinfo{person}{Dezhen Chen}.} \bibinfo{year}{2020}\natexlab{}.
\newblock \bibinfo{booktitle}{\emph{{Chapter 6 - Optimal design of heat
  exchanger networks}}}.
\newblock \bibinfo{publisher}{Academic Press}, \bibinfo{pages}{231--317}.
\newblock


\bibitem[Sporici et~al\mbox{.}(2020)]%
        {sporici2020}
\bibfield{author}{\bibinfo{person}{Dan Sporici}, \bibinfo{person}{Elena
  Cușnir}, {and} \bibinfo{person}{Costin-Anton Boiangiu}.}
  \bibinfo{year}{2020}\natexlab{}.
\newblock \showarticletitle{{Improving the Accuracy of Tesseract 4.0 OCR Engine
  Using Convolution-Based Preprocessing}}.
\newblock \bibinfo{journal}{\emph{{Symmetry}}}  \bibinfo{volume}{12}
  (\bibinfo{date}{05} \bibinfo{year}{2020}), \bibinfo{pages}{715}.
\newblock
\urldef\tempurl%
\url{https://doi.org/10.3390/sym12050715}
\showDOI{\tempurl}


\bibitem[Su et~al\mbox{.}(2021)]%
        {su2021}
\bibfield{author}{\bibinfo{person}{Zhidong Su}, \bibinfo{person}{Weihua Sheng},
  {and} \bibinfo{person}{Senlin Zhang}.} \bibinfo{year}{2021}\natexlab{}.
\newblock \showarticletitle{{Convolutional Neural Network Optimization Using
  Modified NSGA-II}}. In \bibinfo{booktitle}{\emph{2021 IEEE 11th Annual
  International Conference on CYBER Technology in Automation, Control, and
  Intelligent Systems (CYBER)}}. \bibinfo{publisher}{{IEEE Computer Society}},
  \bibinfo{pages}{466--471}.
\newblock
\urldef\tempurl%
\url{https://doi.org/10.1109/CYBER53097.2021.9588230}
\showDOI{\tempurl}


\bibitem[Sulaiman et~al\mbox{.}(2019)]%
        {sulaiman2019}
\bibfield{author}{\bibinfo{person}{Alaa Sulaiman}, \bibinfo{person}{Khairuddin
  Omar}, {and} \bibinfo{person}{Mohammad~Faidzul Nasrudin}.}
  \bibinfo{year}{2019}\natexlab{}.
\newblock \showarticletitle{{Degraded Historical Document Binarization: A
  Review on Issues, Challenges, Techniques, and Future Directions}}.
\newblock \bibinfo{journal}{\emph{{Journal of Imaging}}}  \bibinfo{volume}{5}
  (\bibinfo{date}{04} \bibinfo{year}{2019}).
\newblock
\urldef\tempurl%
\url{https://doi.org/10.3390/jimaging5040048}
\showDOI{\tempurl}


\bibitem[Tensmeyer and Martinez(2020)]%
        {tensmeyer2020}
\bibfield{author}{\bibinfo{person}{Chris Tensmeyer} {and} \bibinfo{person}{Tony
  Martinez}.} \bibinfo{year}{2020}\natexlab{}.
\newblock \showarticletitle{{Historical Document Image Binarization: A
  Review}}.
\newblock \bibinfo{journal}{\emph{{SN Computer Science}}}  \bibinfo{volume}{1}
  (\bibinfo{date}{05} \bibinfo{year}{2020}).
\newblock
\urldef\tempurl%
\url{https://doi.org/10.1007/s42979-020-00176-1}
\showDOI{\tempurl}


\bibitem[Tong and Evans(1996)]%
        {Tong1996}
\bibfield{author}{\bibinfo{person}{Xiang Tong} {and} \bibinfo{person}{David~A.
  Evans}.} \bibinfo{year}{1996}\natexlab{}.
\newblock \showarticletitle{{A Statistical Approach to Automatic OCR Error
  Correction in Context}}. In \bibinfo{booktitle}{\emph{Fourth Workshop on Very
  Large Corpora, VLC@COLING 1996}}. \bibinfo{publisher}{Association for
  Computational Linguistics}, \bibinfo{address}{Copenhagen, Denmark},
  \bibinfo{pages}{88--100}.
\newblock
\urldef\tempurl%
\url{http://www.aclweb.org/anthology/W/W96/W96-0108.pdf}
\showURL{%
\tempurl}


\bibitem[Walker et~al\mbox{.}(2018)]%
        {walker2018}
\bibfield{author}{\bibinfo{person}{Jake Walker}, \bibinfo{person}{Yasuhisa
  Fujii}, {and} \bibinfo{person}{Ashok~C. Popat}.}
  \bibinfo{year}{2018}\natexlab{}.
\newblock \showarticletitle{{A Web-Based OCR Service for Documents}}. In
  \bibinfo{booktitle}{\emph{{13th IAPR International Workshop on Document
  Analysis Systems (DAS)}}}, Vol.~\bibinfo{volume}{1}.
  \bibinfo{address}{Vienna, Austria}, \bibinfo{pages}{21--22}.
\newblock


\bibitem[Weil(2021)]%
        {weil2021}
\bibfield{author}{\bibinfo{person}{Stefan Weil}.}
  \bibinfo{year}{2021}\natexlab{}.
\newblock \bibinfo{title}{{Tesseract-OCR/langdata: Source training data for
  Tesseract for lots of Languages}}.
\newblock
\newblock
\urldef\tempurl%
\url{https://github.com/tesseract-ocr/langdata}
\showURL{%
\tempurl}


\bibitem[Yahui et~al\mbox{.}(2020)]%
        {wang2020}
\bibfield{author}{\bibinfo{person}{Wang Yahui}, \bibinfo{person}{Shi Ling},
  \bibinfo{person}{Zhang Cai}, \bibinfo{person}{Fu Liuqiang}, {and}
  \bibinfo{person}{Jin Xiangjie}.} \bibinfo{year}{2020}\natexlab{}.
\newblock \showarticletitle{{NSGA-II algorithm and application for
  multi-objective flexible workshop scheduling}}.
\newblock \bibinfo{journal}{\emph{Journal of Algorithms \& Computational
  Technology}}  \bibinfo{volume}{14} (\bibinfo{year}{2020}),
  \bibinfo{pages}{1748302620942467}.
\newblock
\urldef\tempurl%
\url{https://doi.org/10.1177/1748302620942467}
\showDOI{\tempurl}


\bibitem[Yang(2021)]%
        {yang2021}
\bibfield{author}{\bibinfo{person}{Xin-She Yang}.}
  \bibinfo{year}{2021}\natexlab{}.
\newblock \bibinfo{booktitle}{\emph{{Chapter 6 - Genetic Algorithms}}
  (\bibinfo{edition}{second edition} ed.)}.
\newblock \bibinfo{publisher}{Academic Press}, \bibinfo{pages}{91--100}.
\newblock
\urldef\tempurl%
\url{https://doi.org/10.1016/B978-0-12-821986-7.00013-5}
\showDOI{\tempurl}


\bibitem[Yu et~al\mbox{.}(1998)]%
        {yu1998}
\bibfield{author}{\bibinfo{person}{M. Yu}, \bibinfo{person}{Nawapak Eua-anant},
  \bibinfo{person}{A. Saudagar}, {and} \bibinfo{person}{Lalita Udpa}.}
  \bibinfo{year}{1998}\natexlab{}.
\newblock \showarticletitle{{Genetic algorithm approach to image segmentation
  using morphological operations}}. In \bibinfo{booktitle}{\emph{{1998
  International Conference on Image Processing. ICIP98 (Cat. No.98CB36269)}}},
  Vol.~\bibinfo{volume}{3}. \bibinfo{publisher}{{IEEE} Computer Society},
  \bibinfo{address}{Chicago, Illinois, USA}, \bibinfo{pages}{775--779}.
\newblock
\urldef\tempurl%
\url{https://doi.org/10.1109/ICIP.1998.999063}
\showDOI{\tempurl}


\end{thebibliography}

\end{document}